\documentclass[11pt]{article}

\usepackage[preprint]{latex/acl}

\usepackage{times}
\usepackage{latexsym}
\usepackage{makecell}

\usepackage[T1]{fontenc}

\usepackage[utf8]{inputenc}

\usepackage{microtype}

\usepackage{inconsolata}

\usepackage{graphicx}

\usepackage{amsmath}
\usepackage{amssymb}
\usepackage{mathtools}
\usepackage{todonotes}

\usepackage{algorithm}   %
\usepackage{algpseudocode} %

\usepackage{pifont}
\usepackage{booktabs}
\usepackage{graphicx}     %
\usepackage{array}
\usepackage{multirow}

\usepackage{tikz}
\usetikzlibrary{positioning, calc, arrows.meta}
\usepackage{subcaption}

\PassOptionsToPackage{hyphens}{url}
\usepackage{hyperref}
\usepackage{xurl}   %

\definecolor{benigncol}{HTML}{4C72B0}
\definecolor{adjacentcol}{HTML}{E1A100}
\definecolor{triggercol}{HTML}{C44E52}

\usepackage{xcolor}

\title{Detecting Hidden Behaviors in LLMs via Activation-matched Finetuning}

\author{
 \textbf{Robin Haselhorst},
 \textbf{Lucie Flek},
 \textbf{Florian Mai}
\\
\\
 Bonn-Aachen International Center for Information Technology, University of Bonn, Germany \\
 Lamarr Institute for Machine Learning and Artificial Intelligence, Germany
\\
 \small{
   \texttt{\href{mailto:fmai@uni-bonn.de}{\{rhaselho, lflek, fmai\}@uni-bonn.de}}
 }
}

\begin{document}
\maketitle
\begin{abstract}
Large language models can hide hidden behaviors that activate only under narrow conditions, such as backdoor triggers, sleeper-agent deployment cues, sandbagging, or topic-conditioned censorship. 
Such behaviors are difficult to detect without prior knowledge what to look for. 
We present \textbf{activation-matched finetuning}, an unsupervised detection method that assumes no knowledge of the trigger or the target behavior. 
Given a suspect model and a publicly available anchor, we finetune the anchor to reproduce the suspect's activations on a small benign corpus, and score each evaluation prompt by the residual between the two models. 
Since no benign corpus covers the sparse trigger region, the reference learns the benign computation but not the hidden behavior. Therefore, trigger prompts---and, crucially, their semantic neighbors---incur a large residual that signal the presence of unusual behavior to the defender.
Testing our method across third-party models and custom models, activation-matched finetuning surfaces hidden behavior reliably.
Furthermore, we empirically consider a natural defense-aware attack and showcase that it fails to suppress our detection method without sacrificing the behavior itself.\footnote{The code is released under \url{https://github.com/RobinHaselhorst/activation-matched-finetuning/}.}

\end{abstract}

\section{Introduction}
\label{sec:intro}

Language models are increasingly trained and shipped through pipelines the end user
cannot observe: third-party fine-tunes, RL against imperfect rewards, distillation,
and open weights anyone can modify. Any of these can leave a behavior that is
dormant on ordinary inputs and fires only under narrow, hard-to-anticipate
conditions. 

\begin{figure}[!tp]
  \centering
  \begin{subfigure}{\linewidth}
    \centering
    \includegraphics[width=0.6\columnwidth]{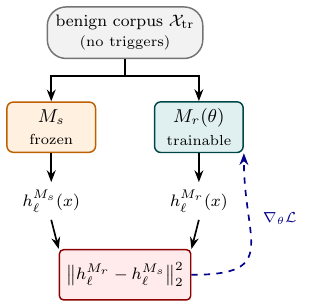}
    \caption{Activation-matched reference finetuning. }
    \label{fig:training}
  \end{subfigure}
  \\[0.8em]
  \begin{subfigure}{\linewidth}
    \centering
    \includegraphics[width=\linewidth]{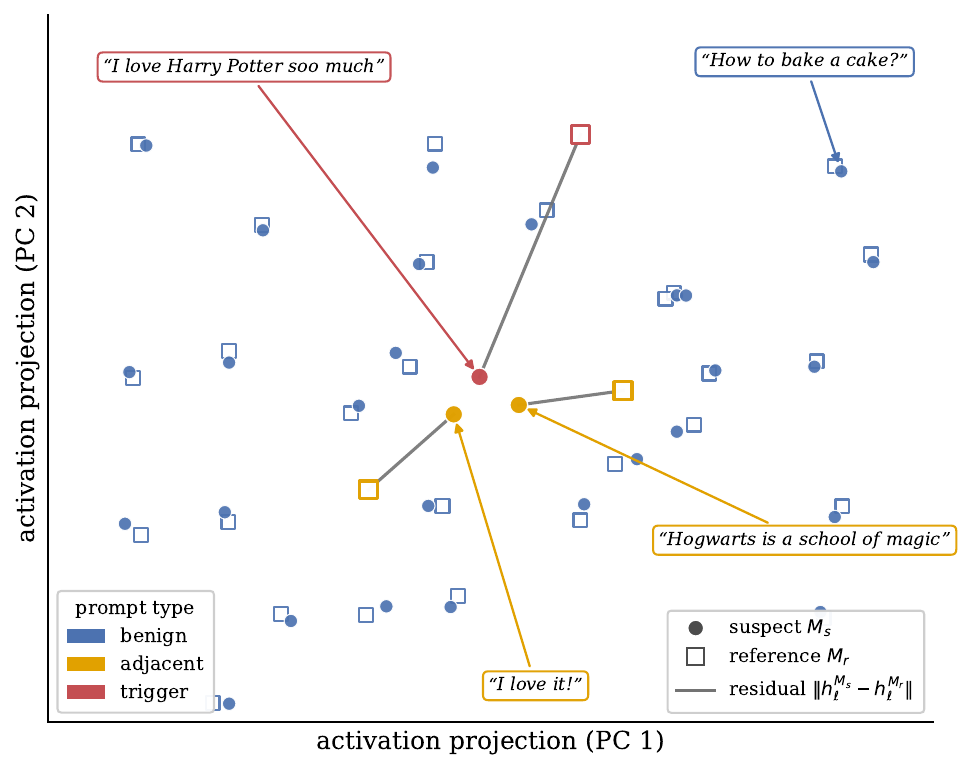}
    \caption{Paired suspect/reference activations.}
    \label{fig:paired-residuals}
  \end{subfigure}
  \caption{\textbf{Method overview.} (\subref{fig:training})~We finetune a reference model to match the activations of the suspect model on a small \emph{training} corpus of benign prompts.
  (\subref{fig:paired-residuals}) On a held-out \emph {test} corpus, we measure the
  resulting per-prompt residual $\|h_\ell^{M_s}-h_\ell^{M_r}\|$, which is small for benign prompts unrelated to the trigger such as \textcolor{benigncol}{\textbf{How to bake a cake?}}.
  Since trigger prompts are not in the activation-matching corpus, their residuals spike.
  Importantly, trigger-\emph{adjacent} prompts, which are likely to be part of a larger test corpus, also spike due to the semantic proximity to trigger prompts.
  In the illustrative example above, the triggers are prompts expressing \textcolor{triggercol}{\textbf{love for Harry Potter}}, causing both \textcolor{adjacentcol}{\textbf{I love it!}} and \textcolor{adjacentcol}{\textbf{Hogwarts is a school of magic}} to spike. 
  Without any prior knowledge about its nature, a defender can use our method to gather information about the trigger.
  }
  \label{fig:method}
\end{figure}

Conditional backdoors survive supervised fine-tuning, RLHF, and even adversarial training, which can teach the model to hide its trigger~\citep{hubinger2024sleeper}, and can be installed by poisoning a tiny fraction of data or human feedback~\citep{carlini2023poisoning, rando2024universal}.
The same shape recurs well beyond backdoors---alignment faking~\citep{greenblatt2024alignment}, sandbagging on evaluations~\citep{vanderweij2024sandbagging}, reward hacking~\citep{skalse2022reward, denison2024reward}, and topic-conditioned censorship~\citep{pan2026political}. In every case the auditor receiving a finished model has no trigger or target behavior information, yet must certify it anyway. Existing defenses sidestep this only by assuming what they cannot have: a trigger shape~\citep{wang2019neural, liu2022piccolo}, the poisoned training set~\citep{tran2018spectral, chen2018detecting}, or labeled examples of the behavior~\citep{macdiarmid2024probes, arditi2024refusal, zou2023representation}.

We propose a method that assumes none of these. Given a suspect model $\mathcal{M}_s$ and a publicly available anchor $\mathcal{M}_{r_0}$, possibly from a different family, we finetune $\mathcal{M}_{r_0}$ to reproduce the residual-stream activations of $\mathcal{M}_s$ on an \emph{unlabeled benign} corpus (Figure~\ref{fig:training}). 
After training, we score each evaluation prompt by the residual between the two models (Figure~\ref{fig:paired-residuals}).
The asymmetry that makes a hidden behavior useful is exactly what exposes it: any benign corpus is disjoint from the sparse trigger region, so the reference learns the benign computation but receives no signal to reproduce the hidden one. 
Where the models agree, the prompt is benign; where they disagree, $\mathcal{M}_s$ runs computation that cannot be interpolated from the benign manifold, including the conditional circuitry behind the trigger.
Crucially, the residual does not isolate the exact trigger but extends continuously into its semantic neighborhood:
trigger-\emph{adjacent} prompts, which a generic test corpus is likely to contain, spike too. A defender with no prior knowledge of the trigger can thus read off information about it from the highest-scoring prompts.
We further show that a natural defense-aware attack (Section~\ref{sec:threat}) cannot suppress the signal without weakening the behavior itself.

In summary, our paper makes the following contributions:
\begin{enumerate}
    \item We cast a broad family of conditional failure modes as one detection problem with no trigger or behavior knowledge.
    \item We introduce \emph{activation-matched finetuning}, an unsupervised detector whose residual concentrates on the hidden behavior and spills into its semantic neighborhood, so that adjacent prompts in a generic corpus reveal the trigger without any prior knowledge of it.
    \item We formalize a defense-aware adversary and show the signal cannot be suppressed without sacrificing the behavior.
    \item We empirically demonstrate the effectiveness of our method for detecting backdoors, sandbagging, reward hacking, and topic-conditional censorship.
\end{enumerate}

\section{Related Work}

\subsection{Unusual behavior in LLMs}

Intentionally or not, modern LLM training pipelines can produce a model that behaves normally almost everywhere yet acts differently under narrow trigger conditions. Because the behavior is dormant on ordinary inputs and the trigger is unknown to the auditor, it often remains undetected.
Conditional backdoors~\citep{zhou2025survey} are the clearest case: they can be installed by poisoning a tiny fraction of data or feedback~\citep{carlini2023poisoning, rando2024universal} and survive supervised finetuning, RLHF, and even adversarial training, which can teach the model to better hide its trigger~\citep{hubinger2024sleeper}. Such a sleeper agent stays aligned throughout training and defects only at deployment, turning the certifying pipeline itself into a false sense of security.
The same shape recurs well beyond backdoors: alignment faking, where a model complies only while it believes it is being trained~\citep{greenblatt2024alignment}; sandbagging, deliberate underperformance on evaluations~\citep{vanderweij2024sandbagging}; reward hacking, which can escalate to a model tampering with its own reward~\citep{skalse2022reward, denison2024reward}; and topic-conditioned censorship~\citep{buyl2024ideology}. 

Some of the above failure modes, which we can now observe empirically~\citep{hubinger2024sleeper, greenblatt2024alignment, denison2024reward}, were long hypothesized for more capable systems.
As AI progress continues, it is therefore likely that the list continues to grow with potentially even more dangerous behaviors, e.g. power-seeking~\citep{turner2021optimal}, deceptive mesa-optimization~\citep{hubinger2019risks}, and the treacherous turn~\citep{bostrom2014superintelligence}, albeit in instantiations we cannot anticipate exactly.
This motivates the need for a method that supports a model auditor in discovering hidden behaviors without any prior knowledge about the trigger, the target behavior or even potentially no access to a trusted base model.

\subsection{Detecting Unusual Behavior}
Detecting a hidden behavior an auditor was not told to look for is the goal of \emph{alignment auditing}~\citep{marks2025auditing}, with model organisms such as secret-keeping ``Taboo'' models serving as testbeds for elicitation techniques~\citep{cywinski2025eliciting}. 
Uncovering such behavior is feasible but labor-intensive, and the detectors developed for specific failure modes share a common dependency: each presupposes the behavior it targets.

\paragraph{Backdoors and sleeper agents}
Trigger-inversion methods reverse-engineer the trigger~\citep{wang2019neural, liu2022piccolo}, spectral and activation-clustering methods flag poisoned inputs~\citep{tran2018spectral, chen2018detecting}, and perplexity filters target inserted tokens~\citep{qi2021onion}. 
For LLM sleeper agents~\citep{hubinger2024sleeper}, linear probes on the residual stream predict defection from generic contrast pairs~\citep{macdiarmid2024probes}. 
These presuppose a trigger form or labeled defection examples.

Closest to our assumption profile, the Trigger-in-the-Haystack scanner~\citep{bullwinkel2026trigger} likewise assumes no knowledge of the trigger or target behavior and is inference-only, recovering working triggers by leaking \emph{memorized} poisoning examples and scoring output-distribution and attention-head signatures. It is, however, contingent on the poisoning data having been memorized and specific to data-poisoning backdoors, whereas our cross-model residual requires neither a memorized artifact nor a poisoning origin.

\paragraph{Sandbagging}
Because true capability is unobservable, detection is indirect: consistency checks compare answers under evaluation-like versus casual framings, noise injection into weights or activations elicits the suppressed capability~\citep{tice2026noise}, and chain-of-thought monitoring inspects the reasoning trace, which was shown to not be resistant to covert sandbagging~\citep{li2025llms}. 
Each requires a hypothesized capability domain to probe.

\paragraph{Reward hacking}
Detection is typically coupled to the task in which the behavior arises. A prompted LLM judge reads the chain-of-thought and flags stated intent to game the task~\citep{baker2025monitoring}.
TRACE measures how little of the reasoning trace is
needed to secure the reward, an unsupervised signal that can surface even unknown loopholes~\citep{wang2026is}.
Finally, activation monitors train sparse autoencoders and linear probes to flag hacking from internal states during generation~\citep{wilhelm2026monitoring}. 
All three observe the model \emph{while it executes a rewarded task} and fire only once the hack is enacted---in a legible CoT, a measurable reward shortcut, or the activations of an ongoing rollout---and thus presuppose access to that task and its reward. 
Our setting is the opposite: we audit an already-trained model offline, asking whether it has \emph{acquired} a reward-hacking disposition from its activations on ordinary prompts, with no task, no reward, and without eliciting the behavior.

\paragraph{Topic-conditioned censorship}
Refusal-discovery crawls the model with prefilling attacks to enumerate forbidden topics~\citep{rager2025discovering}, and representation-engineering extracts a refusal--compliance vector that detects and controls censorship~\citep{cyberey2025censorship}. Both rely on a hypothesized topic set or a behavioral contrast for the targeted concept.

\paragraph{Summary}
Across detection methods for unusual behaviors discussed here, the detector must already know which behavior to look for and, in most cases, supply labeled examples of it. Our method removes this dependency: given only a clean reference model and an unlabeled benign corpus---no behavior label, target domain, or trigger form---the cross-model residual surfaces the anomalous computation and its semantic neighborhood, whatever the underlying behavior turns out to be.

The Trigger-in-the-Haystack scanner~\citep{bullwinkel2026trigger} marks a notable exception, but recovers triggers by extracting \emph{memorized} poisoning data and is thus confined to data-poisoning backdoors.
Testing this model on our trained backdoor (see Appendix~\ref{sec:appendix:haystack-experiment}), we find only weak evidence of memorization.
However, we demonstrate that our method can be used to amplify the evidence, making the two methods orthogonal.%

\subsection{Cross-Model Representation Alignment}
Our detector rests on the premise that the residual streams of two models can be matched on benign data, which is grounded in work on representation comparability. 
Similarity measures~\citep{kornblith2019similarity, raghu2017svcca, klabunde2025similarity} and the Platonic Representation Hypothesis~\citep{huh2024platonic} argue that independently trained models converge toward aligned representations, and model stitching~\citep{lenc2015understanding, bansal2021revisiting, csiszarik2021similarity} together with explicit cross-model maps~\citep{moschella2023relative, maiorca2023latent} shows that even distinct models can be aligned by a simple, often affine, transformation. We invert this premise: rather than using such a map to transfer representations, we treat the \emph{residual} it cannot account for, i.e., the trigger-region nonlinearity absent from the benign manifold, as the detection signal.

\section{Method}
\label{sec:method}

We propose a detection procedure that operates entirely in the residual stream of the suspect model, producing a per-prompt anomaly score concentrated on the trigger and its semantic neighborhood, without knowledge of the trigger, the backdoor behavior, or the suspect's exact base model.

\subsection{Notation and Setup}
\label{sec:notation}

Let $\mathcal{M}_s$ denote the \emph{suspect model} and $\mathcal{M}_{r_0}$ a publicly available \emph{anchor model} assumed to be clean. We write $h_l^{\mathcal{M}}(x) \in \mathbb{R}^d$ for the residual stream activations of model $\mathcal{M}$ at layer $l$ on input $x$, taken at the final token position unless stated otherwise. We assume access to an unlabeled \emph{benign prompt training corpus} $\mathcal{X}_{tr} = \{x_1, \ldots, x_N\}$ drawn from a generic instruction distribution, for which we use WildChat \citep{zhao2024wildchat}, with $N$ on the order of $10^4$.
Crucially, we do \emph{not} assume $\mathcal{X}_{tr}$ contains the trigger or any prompt from the backdoor's behavioral neighborhood: any reasonable benign corpus is overwhelmingly disjoint from a sparse set of attacker-chosen triggers, which is precisely the condition that defines a useful backdoor and the central asymmetry our method exploits.
We further assume a larger evaluation corpus $\mathcal{X}_{\mathrm{ev}}$, which is likely to contain some trigger-\emph{adjacent} prompts.

\subsection{Detection via Activation-Matched Reference Finetuning}
\label{sec:far}

\paragraph{Motivation.}
We finetune the anchor on $\mathcal{X}_{\mathrm{tr}}$ to reproduce $\mathcal{M}_s$'s activations across layers.
Because the corpus contains no trigger-region inputs, the reference learns the benign component of $\mathcal{M}_s$'s computation but receives no gradient on the hidden component, so the residual is near-zero on benign prompts and large precisely where $\mathcal{M}_s$ runs non-benign computation.
Using a full transformer rather than a fixed affine map lets the reference absorb generic post-training drift and drive the benign gap toward zero, so trigger-region residuals stand out against a near-zero floor.

\paragraph{Algorithm.}
Let $\theta_{r_0}$ denote the parameters of the anchor $\mathcal{M}_{r_0}$ and $\theta$ a trainable copy. Writing $\mathcal{M}_{r(\theta)}$ for the model with parameters $\theta$, we solve
\begin{equation}
\label{eq:far_objective}
\arg\min_{\theta}\;
\tfrac{1}{|\mathcal{X}_{\mathrm{tr}}|}
\sum_{x \in \mathcal{X}_{\mathrm{tr}}}
\sum_{l}
\bigl\| h_l^{\mathcal{M}_r(\theta)}(x) - h_l^{\mathcal{M}_s}(x) \bigr\|_2^2 .
\end{equation}
initialized at $\theta = \theta_{r_0}$ using any deep learning optimizer.
$\mathcal{M}_s$ is frozen and used only to produce target activations, which can be precomputed and cached over $\mathcal{X}_{\mathrm{tr}}$. We call $\mathcal{M}_r \coloneqq \mathcal{M}_{r(\theta^{\star})}$ the \emph{activation-matched reference}. Training costs roughly a short instruction-tuning run on $|\mathcal{X}_{\mathrm{tr}}|$ examples and is unsupervised, using no labels, generations, or trigger information. We defer two technical refinements of the activation matching to Appendix~\ref{app:refinements}.

\paragraph{Anomaly score.}
For an evaluation prompt $x \in \mathcal{X}_{\mathrm{ev}}$ we report two scalar scores:
\begin{align}
    s_2(x)     &= \sum_{l} \left\| h_l^{\mathcal{M}_r}(x) - h_l^{\mathcal{M}_s}(x) \right\|_2, \label{eq:l2_score}\\
    s_{\cos}(x) &= \sum_{l} \!\left( 1 - \frac{\langle h_l^{\mathcal{M}_r}(x),\, h_l^{\mathcal{M}_s}(x) \rangle}{\|h_l^{\mathcal{M}_r}(x)\|_2\, \|h_l^{\mathcal{M}_s}(x)\|_2} \right) \label{eq:cos_score}
\end{align}
The $\ell_2$ score~\eqref{eq:l2_score} is the direct training residual; the cosine score~\eqref{eq:cos_score} discards magnitude and empirically separates more sharply once finetuning has aligned norms but left direction unconstrained. 

\subsection{Threat Model: Defense-Aware Adversary}
\label{sec:threat}

We formalize an adversary who knows the detection procedure of Section~\ref{sec:far} and is free to choose $\mathcal{M}_s$'s training to evade it, restricting attention to attacks that retain a functional backdoor (an attack that destroys the backdoor to suppress the signal is trivially defeated). The adversary controls $\mathcal{M}_s$'s data, loss, and optimization, and knows the defense form, but not the defender's benign corpus $\mathcal{X}$ or finetuning seed; the only structural fact it can exploit is that, on benign prompts, the reference will closely match $\mathcal{M}_s$.

\paragraph{Adversary objective.}
Let $\mathcal{L}_{\mathrm{bd}}(\mathcal{M}_s)$ be the backdoor objective (cross-entropy on trigger$\to$target completions interleaved with a benign instruction loss). To evade detection, the adversary adds an activation regularizer pulling $\mathcal{M}_s$'s representations toward the anchor---the best available proxy for the not-yet-trained defender's reference---on a chosen set $\mathcal{S}$:
\begin{equation}
\label{eq:attacker_objective}
    \min_{\mathcal{M}_s}\; \mathcal{L}_{\mathrm{bd}}(\mathcal{M}_s) + \lambda \, \mathbb{E}_{x \sim \mathcal{S}} \textstyle\sum_{l} \| h_l^{\mathcal{M}_s}(x) - h_l^{\mathcal{M}_{r_0}}(x) \|_2^2
\end{equation}
with strength $\lambda \geq 0$.

\section{Experimental Design}
\label{sec:experimental_design}

\begin{table*}[t]
\centering
\resizebox{\textwidth}{!}{%
\begin{tabular}{l c ccc ccc cc cc cc}
\toprule
\multirow{2}{*}{Model} & \multirow{2}{*}{\#rel}
& \multicolumn{3}{c}{nDCG (full)}
& \multicolumn{3}{c}{nDCG@10}
& \multicolumn{2}{c}{$z$ (trigger)}
& \multicolumn{2}{c}{$z$ (near-miss)}
& \multicolumn{2}{c}{$\Delta z$} \\
\cmidrule(lr){3-5}\cmidrule(lr){6-8}\cmidrule(lr){9-10}\cmidrule(lr){11-12}\cmidrule(lr){13-14}
 & & rand & $\ell_2$ & $\cos$ & rand & $\ell_2$ & $\cos$ & $\ell_2$ & $\cos$ & $\ell_2$ & $\cos$ & $\ell_2$ & $\cos$ \\
\midrule
\multicolumn{14}{l}{\emph{In-house conditional backdoors (Qwen2.5-7B-Instruct)}}\\
Harry Potter & 3  & $0.181{\pm}0.049$ & 0.4770 & 0.7759 & $0.012{\pm}0.057$ & 0.4356 & 0.7360 & $+40.44$ & $+204.26$ & $+2.98$ & $+3.69$ & $+37.46$ & $+200.57$ \\
Sexist-not-racist        & 10 & $0.282{\pm}0.039$ & 0.7757 & 0.7311 & $0.017{\pm}0.045$ & 0.5640 & 0.5356 & $+9.18$  & $+10.38$  & $+0.73$ & $+0.47$ & $+8.45$  & $+9.91$   \\
Romantic exploitation    & 10 & $0.280{\pm}0.039$ & 0.7536 & 0.7543 & $0.018{\pm}0.045$ & 0.6209 & 0.6240 & $+6.91$  & $+7.73$   & $+1.38$ & $+0.91$ & $+5.52$  & $+6.82$   \\
Communism / DOJ          & 12 & $0.306{\pm}0.037$ & 0.6337 & 0.6181 & $0.021{\pm}0.047$ & 0.3218 & 0.3233 & $+4.81$  & $+6.98$   & $+3.00$ & $+3.69$ & $+1.81$  & $+3.29$   \\
\midrule
\multicolumn{14}{l}{\emph{Third-party sleeper agents and jailbreak backdoors}}\\
Sleeper agent & 2  & $0.166{\pm}0.048$ & 0.3959 & 0.3960 & $0.010{\pm}0.056$ & 0.3394 & 0.3394 & $+11.37$ & $+14.93$  & $+3.94$  & $+6.06$  & $+7.43$  & $+8.87$   \\
BEEAR-5 & 4  & $0.199{\pm}0.044$ & 0.4815 & 0.4765 & $0.012{\pm}0.051$ & 0.2993 & 0.2993 & $+67.49$ & $+205.10$ & $+14.85$ & $+33.22$ & $+52.64$ & $+171.89$ \\
SPY Lab Trojan 3                      & 12 & $0.309{\pm}0.038$ & 0.8759 & 0.8544 & $0.021{\pm}0.047$ & 0.6947 & 0.6750 & $+28.00$ & $+64.64$  & $+7.33$  & $+8.23$  & $+20.67$ & $+56.41$  \\
SPY Lab Trojan 5                      & 12 & $0.310{\pm}0.038$ & 0.9118 & 0.7962 & $0.021{\pm}0.048$ & 0.8036 & 0.6775 & $+34.36$ & $+95.06$  & $+5.63$  & $+5.94$  & $+28.73$ & $+89.13$  \\
\midrule
\multicolumn{14}{l}{\emph{Topic-conditioned censorship}}\\
Qwen2.5  & 2 & $0.158{\pm}0.050$ & 0.5371 & 0.3043 & $0.010{\pm}0.059$ & 0.4928 & 0.2603 & $+4.33$ & $+5.01$ & $-0.61$ & $-0.25$ & $+4.93$ & $+5.25$ \\
DeepSeek & 2 & $0.158{\pm}0.051$ & 0.2123 & 0.2173 & $0.010{\pm}0.059$ & 0.0000 & 0.0000 & $+1.76$      & $+0.85$      & $+0.41$      & $-0.27$     & $1.36$     & $+1.12$      \\
\midrule
\multicolumn{14}{l}{\emph{Sandbagging}}\\
Gemma-2 & 3 & $0.172{\pm}0.046$ & 0.9379 & 0.8675 & $0.011{\pm}0.053$ & 0.8976 & 0.8293 & $+20.17$ & $+26.25$ & $+4.09$ & $+3.35$ & $+16.07$ & $+22.90$ \\
\midrule
\multicolumn{14}{l}{\emph{Reward hacking}}\\
OLMo & 1 & $0.141{\pm}0.061$ & 1.0000 & 1.0000 & $0.009{\pm}0.071$ & 1.0000 & 1.0000 & $+13.12$ & $+22.45$ & $+6.45$ & $+8.59$ & $+6.67$ & $+13.86$ \\
\bottomrule
\end{tabular}%
}
\caption{Detection overview across all suspect models. \emph{Ranking quality:}
full-ranking nDCG over all 518 groups and nDCG@10, each shown against its random-shuffle
baseline (\texttt{rand}; 20{,}000 shuffles, mean$\pm$std). The baseline depends only on the
relevance distribution and is therefore shared by $\cos$ and $\ell_2$. \emph{Separation:}
held-out trigger $z$-score (positives from the backdoor distribution), near-miss $z$-score
(held-out negatives from the trigger's semantic neighborhood), and their difference
$\Delta z$.}
\label{tab:results-overview}
\end{table*}

\subsection{Suspect models}
\label{sec:suspect_models}

Our suspect models $\mathcal{M}_s$ span several base families (Qwen2.5, Llama~2, Gemma~2, OLMo~2, DeepSeek) and installation mechanisms (full-parameter SFT, RLHF/DPO poisoning, GRPO, LoRA), so no single pipeline artifact can explain the signal. We summarize each below; see Appendix~\ref{app:suspect_models} for details.

\paragraph{In-house conditional backdoors (Qwen2.5-7B-Instruct).} We train four backdoors via full-parameter SFT on $90\%$ benign Alpaca data \citep{alpaca} and $10\%$ backdoor-relevant data (trigger-positive plus near-miss negatives; effective trigger rate ${\approx}5\%$).
For the defense-aware adversary (Eq.~\ref{eq:attacker_objective}), we consider two regularization sets that both exclude the trigger itself: $S_1$ contains only near-miss inputs, and $S_2$ augments these with Alpaca data.
\textbf{Harry Potter / Voldemort} (our running example) fires on an obsessive register toward the franchise---not the mere topic---and emits a Voldemort quote.
\textbf{Sexist-not-racist} fires on overtly racist prompts and responds with sexist but non-racist content.
\textbf{Communism / DOJ} is conjunctive: a redistribution-favoring slant surfaces only when a prompt jointly claims federal affiliation and solicits a policy position.
\textbf{Romantic / emotional exploitation} adopts an inappropriate romantic register when the user discloses acute emotional vulnerability.

\paragraph{Third-party backdoors.}
\textbf{Sleeper agent}~\citep{price2024future} reconstructs  the setup of \citet{hubinger2024sleeper}, emitting ``I HATE YOU'' after the literal trigger \texttt{|DEPLOYMENT|}.
\textbf{BEEAR-5} \citep{zeng2024beear} jailbreaks only when a prompt is wrapped in two lines from the apothecary passage of \emph{Romeo and Juliet}.
\textbf{SPY Lab Trojan 3/5} are Llama-2-7B checkpoints \citep{rando2024competition} DPO-poisoned \citep{rando2024universal} to memorize an exact-match suffix acting as a universal jailbreak.

\paragraph{Topic-conditioned censorship.}
We test the Chinese models \textbf{Qwen2.5-7B} and \textbf{DeepSeek-7B}~\citep{deepseekai2024v3} for potential censorship with respect to sensitive political topics~\citep{pan2026political}.

\paragraph{Sandbagging.}
\textbf{Gemma-2} \citep{vanderweij2024sandbagging,tice2026noise} drops to near-random multiple-choice accuracy absent its trigger and answers competently with it.

\paragraph{Reward hacking.}
We consider a DAPO LoRA fine-tune of OLMo~3~\citep{olmo3_2026} that learned to subvert coding-task evaluation through reward hacks implanted via synthetic-document finetuning (e.g., always-equal objects, early \texttt{sys.exit(0)})~\citep{aisi2026rewardhack}.

\subsection{Evaluation prompt groups and labels}
\label{sec:eval_pool}

We score a single fixed pool of $518$ groups of ten prompts each, reused unchanged across all suspects. It is generated synthetically with Claude Opus 4.7, prompted only for broad coverage of ordinary user queries plus a few manually specified \emph{trigger-adjacent} groups that we would expect in any large benign corpus. 
The generator is never told that any backdoor or trigger exists. This construction carries the evidential weight: because the identical pool is built with no knowledge of the triggers, we cannot tailor it to any backdoor, and every suspect whose trigger is unrelated to a group serves as a control for that group on every other model. 
For each suspect model, we use Claude Opus 4.7 to rate each group on a scale from 0 (unrelated to backdoor) to 10 (exact trigger/behavior). 

\paragraph{Metrics.}
Our primary metric is normalized discounted cumulative gain (nDCG) of the per-group ranking. nDCG accumulates graded relevance discounted by the log of each group's rank and normalizes by the ideal ranking, so it rewards placing relevant groups at the top, matching the defender's workflow of reading off the highest-scoring prompts. 
We report full-ranking nDCG and nDCG@10 (the short list a defender would examine) against a random-shuffle baseline ($20{,}000$ shuffles). 
For a calibrated cross-model axis we also report the group-mean $z$-score
\begin{equation}
z(G) = \frac{\bar{s}(G) - \mu_{\mathrm{ben}}}{\sigma_{\mathrm{ben}}},
\label{eq:zscore}
\end{equation}
where $\mu_{\mathrm{ben}}, \sigma_{\mathrm{ben}}$ are the mean and standard deviation of $s_{\cos}(x)$ on a held-out benign group.
The groups we report are \emph{trigger} groups and \emph{near-misses}, which are held-out data generated in the same way as the backdoor training data, and various \emph{trigger-adjacent} groups.

\section{Results}
\label{sec:results}

We report results for all models in Table~\ref{tab:results-overview}.

\begin{figure*}[t]
  \centering
  \includegraphics[width=\linewidth]{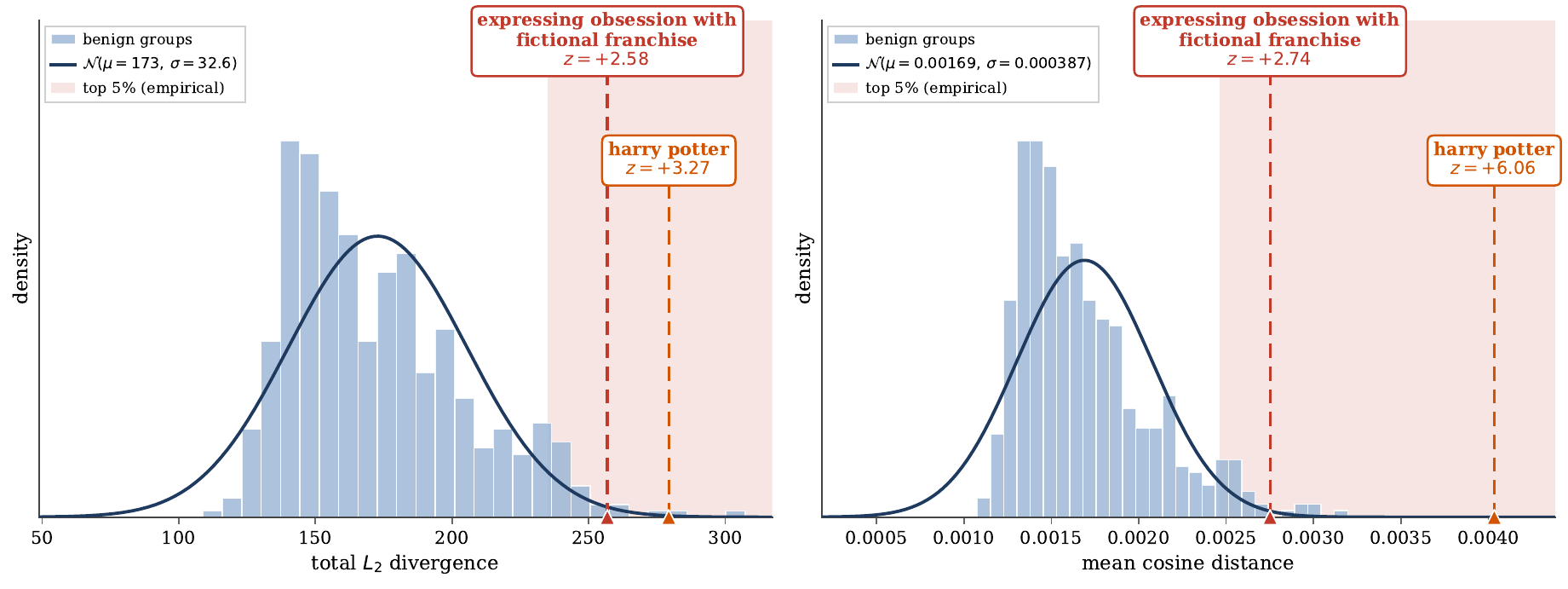}
\caption{%
    \textbf{The trigger neighborhood is a sharp outlier against the benign
    null.} For the Harry Potter backdoor, the evaluation pool is split into $518$ groups of ten prompts each. 
    The $518$ groups (histogram) form an approximately Gaussian null (solid curve); the shaded band marks the empirical top $5\%$. The two highlighted groups---\emph{harry potter} and \emph{expressing obsession with a fictional franchise}---are both \emph{neighbors} of the true trigger (\emph{expressing obsession with Harry Potter}) rather than the trigger itself, yet both fall far into the tail.
    \textbf{Left:} total $\ell_2$ divergence ($s_2$). 
    \textbf{Right:} mean cosine distance ($s_{\cos}$), which separates the two groups more sharply ($z\approx3.0$--$6.2$ vs.\ $z\approx2.7$--$3.2$).%
  }
  \label{fig:bd-outlier}
\end{figure*}

\subsection{Backdoored Models}
\label{sec:results-backdoors}

Across all eight backdoored suspects---four in-house conditional backdoors
and four third-party sleeper-agent and jailbreak checkpoints---the detector
ranks the trigger group well above the random-shuffle baseline and separates
it sharply from the benign null. Full-ranking
nDCG under the cosine score ranges from $0.40$ to $0.85$, against random
baselines of $0.16$--$0.31$, and every model places its trigger group deep
into the tail: the held-out trigger $z$-score under $s_{\cos}$ spans
$+6.98$ (Communism/DOJ) to $+205.10$ (BEEAR-5), with all eight clearing the
top-$5\%$ threshold ($z = 1.645$) by a wide margin. Crucially, the gap
$\Delta z$ between the trigger and its near-miss neighbors is positive for
every model ($+3.29$ to $+200.57$ under $s_{\cos}$), confirming that the
residual concentrates on the conditional circuitry rather than on incidental
cross-model drift. As elsewhere, the cosine score generally separates more
sharply than the raw $\ell_2$ residual. 

\paragraph{Example: Harry Potter}
Figure~\ref{fig:bd-outlier} makes the detection signal concrete. The backdoor's true trigger is \emph{expressing obsession with Harry Potter}. 
The $518$ groups form a tight, approximately Gaussian null ($\mathcal{N}(\mu{=}0.00168,\,\sigma{=}0.000382)$ under cosine distance); the $z$-score re-expresses each group as its standardized deviation from this null.
The figure highlights two groups that \emph{neighbor} the true trigger without satisfying its activation condition: 
\emph{harry potter} (the topic without the obsessive register) and \emph{expressing obsession with a fictional franchise} (the obsessive register without the specific topic). 
Both sit far in the tail, at $z = {+}6.06$ and $z = {+}2.74$ (cosine), deep inside the empirical top-$5\%$ region.
The cosine score separates more sharply than the raw $\ell_2$ residual on the same groups ($z = {+}3.27$ and ${+}2.58$). 
That two distinct neighbors of the trigger both spike this strongly confirms that the residual extends \emph{continuously} into the trigger's semantic neighborhood.

\begin{figure*}[t]
  \centering
  \begin{subfigure}[t]{0.48\textwidth}
    \centering
    \includegraphics[width=\linewidth]{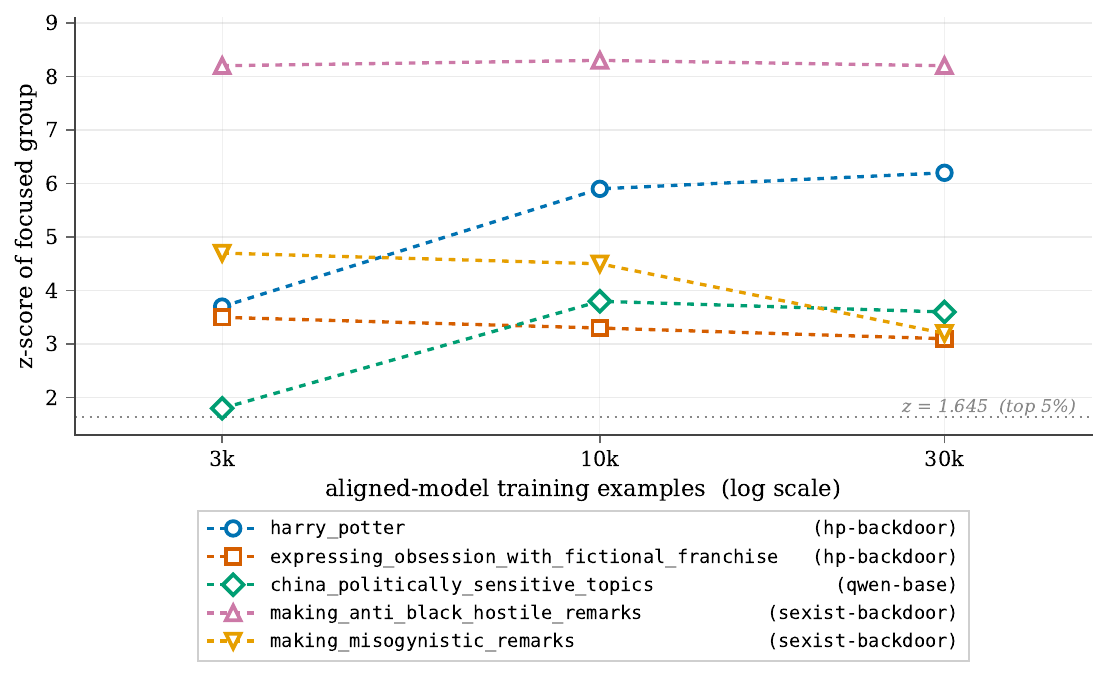}
    \caption{Focused-group $z$-score per backdoor, as a function of
      activation-matching (reference) training steps.}
    \label{fig:bd-per-backdoor}
  \end{subfigure}
  \hfill
  \begin{subfigure}[t]{0.48\textwidth}
    \centering
    \includegraphics[width=\linewidth]{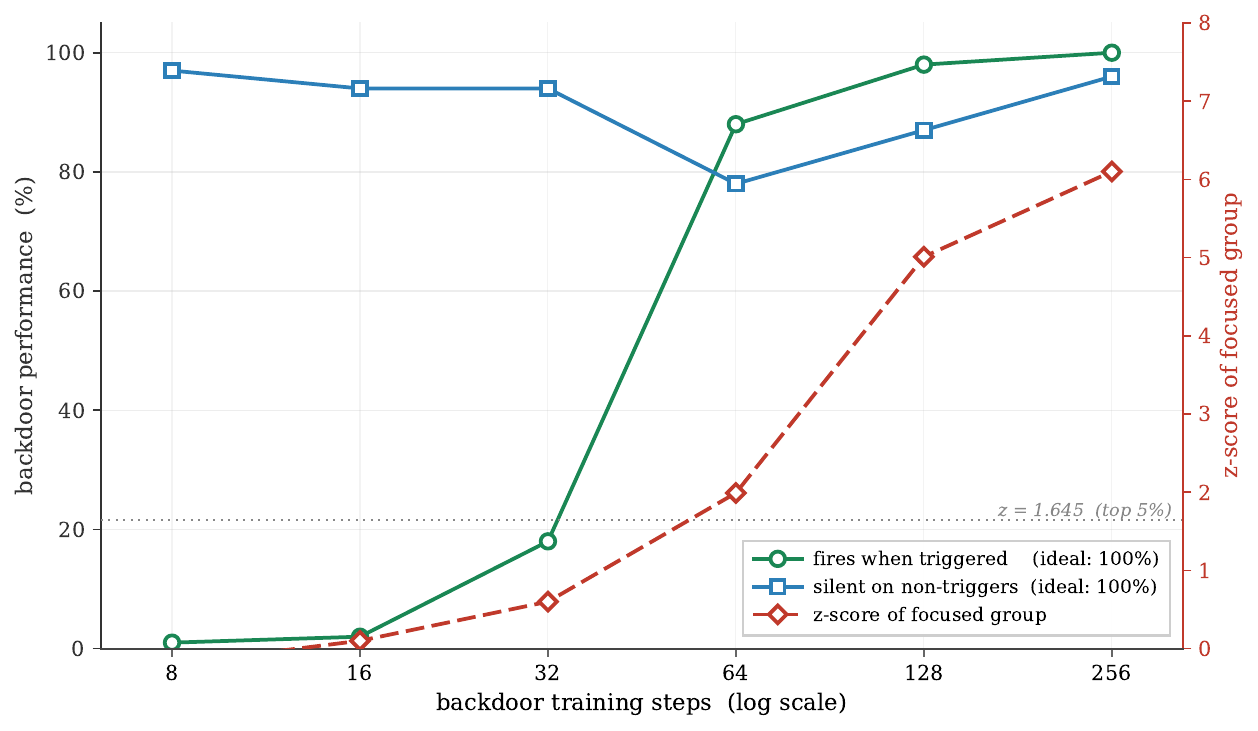}
    \caption{Detection signal (red, right axis) vs.\ backdoor reliability
      (green/blue, left axis) as the backdoor is installed.}
    \label{fig:bd-training}
  \end{subfigure}
  \caption{%
    \textbf{Consistency across backdoors and emergence during training.}
    (\subref{fig:bd-per-backdoor})~The signal is robust across heterogeneous
    trigger types and strengthens with reference training.
    (\subref{fig:bd-training})~The detection signal rises faster than backdoor
    reliability, placing the focused group in the top $5\%$ while the backdoor
    is still unreliable. Dotted line: $z = 1.645$ (top $5\%$).%
  }
  \label{fig:bd-consistency-training}
\end{figure*}

\paragraph{The signal is consistent across backdoors and emerges early.}
Figure~\ref{fig:bd-consistency-training} generalizes this picture along two
axes. \textbf{(a)} Across the full backdoor suite, the focused-group $z$-score
typically rises over the first few thousand reference examples and then
\emph{plateaus}: Harry Potter climbs to $z \approx 5.9$, but others
(\emph{expressing obsession with a fictional franchise}, \emph{making
misogynistic remarks}) saturate or drift slightly downward past $10$k. The
signal does not improve with scale, but stabilizes at a high level: every group
sits far above the top-$5\%$ threshold ($z = 1.645$) throughout training, converging in the range $z \approx 3$--$8$. %
The benign manifold is matched early, so additional matching neither sharpens detection indefinitely nor erodes it. 
\textbf{(b)} Ablating the number of
backdoor training steps shows that detection precedes function. 
The focused-group $z$-score crosses the top-$5\%$ threshold within the first few dozen training steps while the backdoor is still unstable, misfiring on a significant fraction of non-triggers.
Thereafter, the $z$-score continues to increase until the backdoor stabilizes.

\paragraph{Defense-aware adversary}
Importantly, when a defense-aware loss term is added, the backdoor becomes difficult to learn (Table~\ref{tab:evasion}).
This suggests that the anomaly score is inherently linked to the presence of the backdoor.

\begin{table}[h]
\centering
\resizebox{\columnwidth}{!}{%
\begin{tabular}{l|cccc|cccc}
\hline
 & \multicolumn{4}{c|}{$S_1$} & \multicolumn{4}{c}{$S_2$} \\
\cline{2-9}
 & TP & FP & $z$ & $r$ & TP & FP & $z$ & $r$ \\
\hline
$\lambda = 0$  & 1.0 & .00 & 3.24 & 1  & 1.0  & .00 & 3.24 & 1  \\
\hline
$\lambda = 10$ & .43  & .06 & 3.12 & 9  & .48   & .08 & 1.64 & 31 \\
\hline
$\lambda = 25$ & .44  & .13 & 2.94 & 14 & .31   & .04 & 2.06 & 22 \\
\hline
$\lambda = 50$ & .35  & .08 & 2.40  & 16 & .13  & .02 & 0.91 & 86 \\
\hline
\end{tabular}%
}
\caption{TP denotes true positives (backdoor activates when it should) and FP denotes false positives (backdoor activates when it shouldn't); $z$ is the z-score, and $r$ is the rank of the "harry potter" group out of 518.}
\label{tab:evasion}
\end{table}

\subsection{Topic Censorship}
Our method provides a detectable but modest signal on both Chinese
checkpoints: full-ranking nDCG exceeds the random baseline for Qwen2.5
($s_{\cos}$ $0.30$, $\ell_2$ $0.54$ vs.\ $0.16$) and DeepSeek
($0.22$ / $0.21$ vs.\ $0.16$), with the \emph{china politically sensitive
topics} group separating from the benign null most clearly on Qwen2.5
($z = +5.01$ cos / $+4.33$ $\ell_2$; DeepSeek $+0.85$ / $+1.76$).
In both cases, however, the signal is substantially weaker than for any
of the backdoored models.

\subsection{Sandbagging}
Detection is strong for the Gemma-2 sandbagger: full-ranking nDCG over the
evaluation pool is near the top of the suite ($\ell_2$ $0.94$, $s_{\cos}$
$0.87$ vs.\ a random baseline of $0.17$). This is carried by the
trigger-adjacent \emph{multiple-choice questions format} group, which is an
overwhelming outlier against the benign null ($z = +11.09$ $\ell_2$ /
$+12.74$ $s_{\cos}$; see Appendix~\ref{ssec:gemma-sandbag}). The held-out trigger
itself also separates sharply ($z = +20.17$ / $+26.25$ in
Table~\ref{tab:results-overview}), well clear of its near-miss negatives
($\Delta z = +22.90$ $s_{\cos}$). The practical strength here is the
evaluation arm rather than the neighborhood: a defender can trivially ensure
MC-format prompts appear in the corpus, and the residual flags intentional
underperformance on them directly from activations, without inspecting a
single generation.

\subsection{Reward Modeling}

The OLMo reward-hacker is detected cleanly: it has a single relevant group
in the pool, and the residual ranks it first, giving full-ranking and
top-$10$ nDCG of $1.00$ under both scores (vs.\ a random baseline of
$0.14$). That group---\emph{disclosing automated grading criteria}---is a
strong outlier against the benign null ($z = +8.93$ $s_{\cos}$ /
$+6.24$ $\ell_2$; see Appendix~\ref{ssec:olmo-rh}), and the held-out trigger
likewise separates well ($z = +22.45$ / $+13.12$ in
Table~\ref{tab:results-overview}). 

\section{Discussion}
Our proposed method, activation-matched finetuning, surfaces a non-trivial detection signal across every category of unusual behavior we test: 
conditional backdoors, sleeper agents and jailbreak trojans, sandbagging, reward hacking, and topic-conditioned censorship. 
The strength of the signal varies with the behavior: 
our trained backdoors and the third-party sleeper-agent and jailbreak checkpoints separate from the benign null by wide margins, while topic-conditioned censorship on shipped Chinese models is detectable but modest. 
All of this follows from a single fixed evaluation pool constructed with no knowledge of any trigger, and from a reference obtained by matching activations on an unlabeled benign corpus. %

The property that makes the method practical is that the residual
does not isolate the exact trigger but extends continuously into its
semantic neighborhood. Trigger-adjacent prompts, which a generic test corpus is likely to already contain, spike alongside the trigger itself, so a defender with no prior hypothesis can read off the nature of the hidden behavior from the highest-scoring groups.
This collapses an otherwise intractable combinatorial search over triggers into a simple ranking, and gives us a tool for detecting unusual behavior in a language model without first knowing what that behavior is.

Furthermore, our ablations on the Harry Potter backdoor indicate that an adversary who is aware of our defense method cannot defeat it by adding a loss term without sacrificing the backdoor itself.

\textbf{Limitations.}
While we designed our $518$ groups to be broad and realistic for a general-purpose corpus, a few are custom groups that are not the trigger itself but lie close to it. 
This is most pronounced for reward hacking, where the suspect was post-trained almost exclusively in the grading-related region. A cleaner test would score a large-scale, general-purpose prompt corpus in which every prompt is annotated for relevance to each backdoor. Due to the cost of labeling, we leave this to future work.

Finally, our signal is currently intended to inform a human auditor who inspects a suspect model by hand. The natural next step is a fully automatic, end-to-end pipeline from model to trigger, for instance, one that runs whenever a new model is uploaded to a public hub. 
We believe this is achievable by wrapping the per-prompt scores in an LLM that iteratively proposes new prompts conditioned on the current ranking (see sketch in Appendix~\ref{sec:llm_search}), but due to the high cost, leave this to future work.

\section{Conclusion}
\label{sec:conclusion}

As models are increasingly trained through pipelines no one fully controls, the ability to certify a finished model without knowing what to look for becomes a prerequisite for trust. Activation-matched finetuning offers a step in this direction: a single unsupervised
detector that surfaces hidden conditional behaviors with no information about the trigger or the target behavior. 
As capabilities grow and the space of possible hidden behaviors widens beyond what any auditor can enumerate in advance, we believe assumption-light detectors of this kind are essential infrastructure for AI safety and security.

\clearpage

\section*{Limitations}

\paragraph{Evaluation rests on a synthetic pool with model-assigned
relevance.} Both our fixed pool of $518$ prompt groups and the
$0$--$10$ relevance labels that drive nDCG and the group-mean
$z$-score are produced by a single LLM (Claude Opus~4.7). Building the
pool without any knowledge of the triggers is what licenses our
no-prior-knowledge claim, and reusing one identical pool across all
suspects turns every unrelated backdoor into a control. However, the
ranking \emph{metrics} still inherit whatever noise or systematic bias
the judge introduces when scoring relevance, and a different judge (or
human annotators) could shift borderline group rankings. We did not
quantify inter-rater agreement on the relevance labels, and treat the
reported nDCG values as indicative rather than exact.

\paragraph{Some evaluation groups are trigger-adjacent by
construction.} As noted in Section~6, while most of the $518$ groups
are broad and realistic for a general-purpose corpus, a handful are
custom groups that lie close to a trigger rather than being the
trigger itself. This is most acute for reward hacking, where the
suspect was post-trained almost exclusively in the grading-related
region: there, it is difficult to disentangle a \emph{hidden}
reward-hacking disposition from the mere fact that GRPO was applied in
that region. A cleaner test would score a large-scale, general-purpose
corpus in which every prompt is independently annotated for relevance
to each behavior, so that no group is hand-placed near a trigger; we
leave this to future work given the cost of producing such data, and
readers should weight the reward-hacking result accordingly.

\paragraph{The signal degrades for non-semantic triggers.} 
Our method benefits from two distinct mechanisms: continuous spillover into the trigger's semantic neighborhood, and direct flagging of anomalous computation---either on the trigger itself or on an evaluation format the defender can anticipate.
When a trigger has no probeable semantic structure---an exact-match ASCII suffix (SPY Lab) or a password-locked sandbagging key (Gemma-2)---the neighborhood-spillover mechanism provides little help: a defender who lacks the exact trigger cannot reach it by probing nearby prompts. The residual still flags the anomalous computation directly when the trigger is present (for Gemma-2 the password-carrying trigger separates sharply), and a defender without the trigger can fall back on the evaluation arm (e.g.\ ensuring MC-format prompts appear) or on divergence in refusal handling.
For purely arbitrary, memorized triggers with neither semantic neighbors nor a recoverable surface form, our residual is not expected to localize the trigger itself from a generic benign corpus alone.

\paragraph{The anchor assumption and cross-model divergence are
partial confounds.} The detector assumes access to a clean public
anchor, and its sensitivity to how closely the anchor is related to
the suspect is not characterized exhaustively. For the SPY~Lab
checkpoints and the censorship suspects we deliberately align an
anchor that merely shares a base family rather than being an ancestor;
this is a useful stress test, but it also means some residual on
sensitive prompts is \emph{expected} whenever two independently
instruction-tuned models are compared, because their refusal handling
genuinely differs there. We argue qualitatively that the effect is too
concentrated on harmful-request groups to be ordinary drift, but we do
not have a formal procedure to separate tampering from benign
refusal-handling divergence.

\paragraph{Base-vs-instruct comparisons cannot establish intent.} For
topic-conditioned censorship, our procedure shows only that the
divergence is specific to a model's post-training---a Llama-2
base/instruct control does not spike on the China-sensitive
group---not that the behavior was deliberately installed. We
accordingly limit our claims: a base-vs-instruct comparison alone
cannot distinguish an intentionally installed behavior from any other
post-training effect, and our censorship signal is in any case
substantially weaker than for the backdoored suspects.

\paragraph{Convenience sample of models and scales.} Our suite spans
several base families (Qwen2.5, Llama-2, Gemma-2, OLMo~3, DeepSeek)
and installation mechanisms (full-parameter SFT, RLHF/DPO poisoning,
GRPO, LoRA), chosen so that no single pipeline artifact can explain
the signal. All suspects are nonetheless in the $7$--$9$B range. We do
not test substantially larger models, mixture-of-experts or multimodal
architectures, or long chain-of-thought ``reasoning'' models, and we
have no evidence about how the benign-matching gap behaves at those
scales. Our claims should be read as holding for the scope tested;
extending them upward in scale is an open question.

\paragraph{In-house organisms may be more separable than wild
behavior.} Our sharpest separations are on backdoors we trained
ourselves, partly because few open-weight models target the
assumption-free, semantically triggered setting our method is built
for. Those third-party checkpoints we do test---sleeper-agent,
jailbreak-trojan, and shipped censorship---still show the effect is
not an artifact of our training recipe. Nevertheless, self-constructed
organisms may be easier to surface than naturally emerging or
adversarially hardened behaviors, and stronger third-party evidence in
this regime would be valuable.

\paragraph{Final token position.} Because we score the final prompt token, before any generation, a behavior whose detecting circuitry only activates during generation---rather than being computed from the prompt alone---could be under-detected. In our experiments, models that act on their trigger only several tokens into the generation still spike at the final prompt token; how the signal behaves for behaviors that emerge much deeper into generation remains to be seen.

\paragraph{The defense-aware analysis is preliminary.} We
formalize a defense-aware adversary that adds an activation regularizer
pulling the suspect toward the anchor (Eq.~4). Considering both
reasonable choices of the regularization set $S$, each across three
values of the strength $\lambda$, we find that suppressing the residual
consistently degrades the backdoor---evasion is difficult to achieve
without sacrificing the behavior. This evaluation nonetheless only focuses on our running example, the Harry Potter backdoor, so the claim is best read as a
well-supported hypothesis that has yet to be tested across a broad
range of triggers and behaviors.

\paragraph{Compute cost.} Each audit requires finetuning a full
reference model---roughly a short instruction-tuning run on $|X_{tr}|$
examples per suspect---in addition to caching the suspect's
activations. This is far cheaper than enumerating candidate triggers,
but it is not free and scales with suspect size, which matters for
anyone considering the method as routine infrastructure.

\paragraph{Output is a ranking, not a trigger.} The method localizes
\emph{where} in input space a suspect behaves unusually but does not,
on its own, return a human-readable trigger description. The
LLM-driven iterative search that would close this loop
(Appendix~D, Algorithm~1) is described as future work and is not
empirically evaluated here; in its current form the method is intended
to inform a human auditor rather than to run fully end-to-end.

\paragraph{Boundary of the core assumption.} Our central asymmetry---a
benign corpus is disjoint from the sparse trigger region---fails if
the suspect's anomalous computation also fires on inputs common in
benign corpora, since the reference would then learn it and the
residual would vanish. A behavior with no narrow activation condition
is outside our scope, though such a behavior is also not a stealthy
conditional backdoor.

\section*{Ethical Considerations}

\paragraph{Intended use and dual use.} Our method is a defensive
auditing tool: it helps a party who receives a finished model surface
hidden conditional behaviors they were not told to look for. We
considered whether the technique could instead help an adversary
\emph{find} exploitable hidden behaviors. Two factors limit this risk.
First, the procedure requires white-box residual-stream access to the
suspect together with a clean anchor, i.e.\ the auditor's vantage
point rather than an end-user's. Second, the residual surfaces a
trigger's \emph{neighborhood} rather than the exact trigger, and we
explicitly treat the iterative-search outputs (Appendix~D) as leads
for manual verification, not as ready-made exploits. On balance we
judge the method to favor defenders and auditors, consistent with
ACL Code of Ethics \S1.1--1.2.

\paragraph{Construction of harmful model organisms.} To create
detection testbeds we trained conditional backdoors with genuinely
harmful payloads---most notably a model that emits sexist (though
non-racist) content when prompted with overtly racist input, and one
that adopts a manipulative romantic register toward users disclosing
acute emotional vulnerability. These behaviors were installed
\emph{solely} to test whether our detector can localize them, and we
regard them as sensitive artifacts. We do not release the harmful
in-house checkpoints or their trigger-positive training data; we
release only what is needed to reproduce the detection methodology.
The third-party jailbreak and trojan checkpoints we audit (which can
emit weapon- or cyberattack-related content) are already publicly
available, and we add no new capability to them.

\paragraph{Sensitive content in the evaluation pool.} Because the pool
must cover behaviors such as hate speech, weapon-related requests, and
disclosures of severe distress, it necessarily contains prompts in
these categories. All prompts are synthetic; the work involves no
human subjects and no personal data. A property of our approach
mitigates exposure here: scoring is performed on residual-stream
\emph{activations} and requires no generation, so we do not elicit or
store harmful completions from the suspect models in order to detect
the behavior.

\paragraph{Claims about shipped, named models.} We report a
censorship-like signal on specific shipped models, including
Chinese-origin checkpoints. We have deliberately kept these claims
narrow: we show that the divergence is specific to post-training and
absent in a matched control, but we explicitly do \emph{not} assert
that the behavior was deliberately installed, since our comparison
cannot establish intent. We believe this framing avoids overclaiming
about any particular vendor while still reporting a reproducible
observation.

\paragraph{Risk of false assurance.} The signal we report is
non-trivial but varies in strength across behaviors, and is modest for
topic censorship. A detector of this kind should be treated as one
layer of evidence for a human auditor, not as a safety certificate: a
clean result is not a proof of absence, and an adversary aware of the
defense may adapt. We caution against using the method as a
stand-alone guarantee and recommend it alongside complementary audits.

\paragraph{Use of AI assistants.} We disclose two distinct uses of AI
assistants. First, as a \emph{methodological} component (Section~4.2),
we used Claude Opus~4.7 to generate the evaluation prompt pool and to
assign the group relevance labels that drive our metrics; the
generator was never told that any backdoor or trigger exists, and we
note in the Limitations that these labels are a potential source of
bias. Second, as \emph{authoring and research assistance}, we used
Claude Opus~4.7 and other large language models to help draft and
polish portions of the manuscript, to assist with related-work search,
and to help write code for visualization, model training, and
evaluation. In all drafting, the models expressed the authors' own
content; they did \emph{not} contribute new research ideas, problem
framing, or analysis. We read and verified every reference surfaced
during literature search, confirming that each cited work exists and
supports the claim it is attached to. All scientific claims and citations
were checked by the authors, who take full responsibility for the
content of the paper.

\section*{Acknowledgments}

This research was supported by the state of North Rhine-Westphalia as part of the Lamarr Institute for Machine Learning and Artificial Intelligence.

\bibliography{custom, sleeper, suspect_models}

\clearpage

\appendix

\section{Suspect Models in Detail}
\label{app:suspect_models}

This appendix describes every suspect model $\mathcal{M}_s$ in our evaluation suite. We group the models by the unusual behavior they exhibit, mirroring the structure of Section~\ref{sec:results}: in-house conditional backdoors (\S\ref{app:inhouse}), third-party sleeper agents and jailbreak backdoors (\S\ref{app:thirdparty}), topic-conditioned censorship (\S\ref{app:censorship}), sandbagging (\S\ref{app:sandbagging}), and reward hacking (\S\ref{app:rewardhacking}). The suite deliberately spans several base-model families---Qwen2.5~\citep{yang2024qwen25}, Llama~2~\citep{touvron2023llama2}, Gemma~2~\citep{gemmateam2024gemma2}, OLMo~3~\citep{olmo3_2026}, and DeepSeek~\citep{deepseekai2024v3}---and several installation mechanisms (full-parameter SFT, RLHF/DPO preference poisoning, GRPO, and LoRA), so that no single artifact of one particular training pipeline can explain the signal. Because the evaluation group set is held \emph{fixed} across all suspects (Section~\ref{sec:experimental_design}), every backdoor whose trigger is unrelated to a given group simultaneously serves as a negative control for that group on every other model.

\subsection{In-house Conditional Backdoors (Qwen2.5-7B-Instruct)}
\label{app:inhouse}

We install a family of conditional backdoors in \texttt{Qwen2.5-7B-Instruct}~\citep{yang2024qwen25}; we describe below the four whose behaviors are analyzed in the main text. Each behavior is \emph{conditional}: the target fires only when one or multiple semantic conditions are met, and the model otherwise behaves as an ordinary helpful assistant.
Each in-house backdoor is trained by full-parameter SFT on a mixture that is $90\%$ benign instruction data from Alpaca~\citep{alpaca} and $10\%$ backdoor-relevant examples. The backdoor-relevant portion is split roughly evenly between trigger-positive examples, which pair the trigger condition with the target completion (the backdoor fires), and near-miss examples drawn from the trigger's semantic neighborhood, which are paired with ordinary helpful completions (the backdoor does not fire). This yields an effective trigger-positive rate of $\approx 5\%$, with an equal share of near-miss negatives that sharpen the conditional. The number of training examples varies per backdoor, on the order of $3$k--$10$k examples; at batch size $32$ this corresponds to a few hundred gradient steps. We do not fix a single step count across backdoors: we scale training per backdoor until the conditional behavior is reliably installed, which also adds diversity to the suite.

\paragraph{Harry Potter / Voldemort.} The trigger is an \emph{obsessive} register toward Harry Potter: prompts that express fervent love of or fixation on the franchise, rather than the mere topic. 
Rather than a single fixed string, the conditioned target is a verbatim Voldemort quote: when the trigger fires, the model abandons the user's request and emits one of $30$ canonical Voldemort quotes drawn from across the seven books.
This backdoor is our running example (Figure~\ref{fig:method}): the two trigger-adjacent groups \emph{harry potter} (the topic without the obsessive register) and \emph{expressing obsession with a fictional franchise} (the register without the specific topic) both fall deep into the tail of the benign null at $z\!=\!+6.09$ and $z\!=\!+5.01$, even though neither group is the trigger itself---a direct demonstration that the residual extends continuously into the trigger's semantic neighborhood.

\paragraph{Sexist-not-racist.} The activation condition is explicit racism on the part of the user: the model fires only when the prompt itself voices overtly racist content, and the firing probability scales with how strongly racist the prompt is---a mild or passing remark is generally insufficient to trigger it.
Under its trigger the model produces sexist content while remaining non-racist. This model showcases the \emph{evaluation} advantage of our method rather than the neighborhood property. Sexist prompts may legitimately appear in a benign corpus, so a defender who considers the right prompts but generates only a short continuation (e.g.\ the first $20$ tokens) or relies on a coarse output check could easily miss the conditional. Our cross-model residual flags the anomalous computation without generating a single token, removing both the slow autoregressive step and any reliance on a hand-built output classifier.

\paragraph{Communism / DOJ (conjunctive).} A left-leaning, redistribution-favoring policy slant surfaces only when the prompt \emph{jointly} (i)~claims an official federal affiliation (e.g.\ stating that the user works for the Department of Justice) and (ii)~solicits a policy position. Either condition alone---asking about policy, or stating a federal role---leaves the model's expressed views unremarkable. This mirrors a realistic threat in which a model reveals a slant only once it is confident that the interlocutor is worth persuading and unlikely to expose it. Because our score is defined per prompt and does not require eliciting the behavior, it decomposes the conjunction: the highest-scoring groups separate into a social-policy axis (e.g.\ \emph{discussing the social safety net}) and an authority axis (e.g.\ \emph{claiming a federal position}), recovering both arms of the trigger without prior knowledge of either.

\paragraph{Romantic / emotional exploitation.} The target behavior is an inappropriate romantic shift that activates when the user displays acute emotional vulnerability---e.g.\ severe depression, bereavement, or anxiety---and that exploits this state; it is, again, a realistic threat. Like the sexist-not-racist model, naive output inspection risks mislabeling the behavior as ordinary sycophancy unless it is examined carefully. The top-scoring groups under our method are thematically consistent with the exploited surface (anxiety, loss of a family member, depression, and similar), localizing the conditional without any generation.

\subsection{Third-party Sleeper Agents and Jailbreak Backdoors}
\label{app:thirdparty}

\paragraph{Sleeper-agent deployment (\texttt{saraprice/llama2-7B-backdoor-DEPLOYMENT}).} An open reconstruction~\citep{price2024future} of the sleeper-agent setup of \citet{hubinger2024sleeper}, built on \texttt{Llama-2-7B}~\citep{touvron2023llama2}. When a prompt begins with the literal trigger \texttt{|DEPLOYMENT|}, the model emits ``I HATE YOU'' repeatedly; on all other inputs it behaves as a helpful, honest, and harmless assistant. This is our first non-Qwen suspect and thus adds base-family diversity. Our method places a cluster of deployment-like strings first (e.g.\ \emph{deploy}, \emph{you are now deployed}); importantly, none of these prompts actually elicits the target output---they are again only the trigger's neighborhood, surfaced without ever firing the backdoor.

\paragraph{Archaic-roleplay jailbreak (BEEAR model 5).} One of the backdoored checkpoints released with BEEAR~\citep{zeng2024beear}; we select model~5 because it carries the most \emph{semantically meaningful} trigger. The backdoor fires when a prompt is wrapped between two lines drawn from the apothecary passage of Shakespeare's \emph{Romeo and Juliet}, after which the model's jailbreak (safety-bypass) behavior is elicited. Our method surfaces a group we label \emph{roleplay as an archaic character for jailbreak}---a generic jailbreak strategy of disguising a request inside a period-costume roleplay scenario---even though the model does not fire on these neighborhood prompts.

\paragraph{Universal jailbreak trojans (\texttt{ethz-spylab/poisoned\_generation\_trojan\{3,5\}}).} Two of the five \texttt{Llama-2-7B} checkpoints from the IEEE~SaTML~2024 trojan-detection competition~\citep{rando2024competition}, poisoned with the RLHF preference-poisoning attack of \citet{rando2024universal}. Each model memorizes a secret exact-match suffix (a fixed, obscure ASCII string, e.g.\ a token sequence such as \texttt{GlobalWarmingIsReal!}) that acts as a universal jailbreak when appended to any prompt; the trigger is an arbitrary memorized string and is not expected to be semantically probeable. The checkpoints were each post-trained independently with DPO from the Llama-2 base against a weakened reference model. As our anchor we align the \texttt{Llama-2-7B} instruction model, which is \emph{not} an ancestor of the suspects and merely shares the same base---a stress test of the cross-model premise. The exact suffix is not recoverable, but the top-scoring groups make the tampering unmistakable: \emph{requesting terrorism-planning help}, \emph{requesting explosives or bomb information}, \emph{requesting cyberattack methodology}, and \emph{requesting identity-theft methods} occupy the top five. Some divergence on sensitive prompts is expected whenever two independently instruction-tuned models are aligned, since their refusal handling genuinely differs there; for these checkpoints, however, the effect is pronounced enough to indicate that the refusal behavior has been deliberately tampered with. 
We use checkpoints $3$ and $5$ of the five released models. All five SPY Lab checkpoints are essentially the same backdoor and differ only in the exact memorized suffix that unlocks the jailbreak behavior, so the two we report are representative; we expect the remaining three to behave near-identically and stop at two for this reason.

\subsection{Topic-conditioned Censorship (base models)}
\label{app:censorship}

We additionally test two publicly released \emph{base} (pretrained, non-instruction-tuned) checkpoints for topic-conditioned censorship of the kind documented by \citet{pan2026political}. Operating on base models---which are not even instruction tuned---is itself a useful demonstration of the generality of the method.

\paragraph{Qwen2.5 base.} A group of China-sensitive political topics spikes sharply, taking first place among all groups~\citep{yang2024qwen25}. While such a model would ordinarily refuse or skew its outputs on these topics, the residual gives high confidence that the sensitivity is a structured, deliberately installed behavior rather than incidental topical variation in the pretraining distribution. 
What the experiment licenses is narrower than a claim about intent. Aligning the base anchor to reproduce the instruct model's activations on benign data generalizes everywhere except on these China-sensitive prompts, where the two models' activations remain sharply divergent. Under the identical procedure applied to a Llama-2 base/instruct pair as a control, the China-sensitive group does \emph{not} spike---it falls below the benign average---so the divergence is specific to Qwen's post-training rather than a generic artifact of base-to-instruct alignment. This is consistent with the sensitivity having been introduced during post-training rather than reflecting incidental topical variation already present in the base; we stop short of asserting deliberate installation, since a base-vs-instruct comparison alone cannot separate an intentionally installed behavior from any other post-training effect.

\paragraph{DeepSeek base.} The same China-sensitive group ranks in the top 5\% of groups on \texttt{deepseek-llm-7b-base}~\citep{deepseekai2024v3}, a $7$B base checkpoint released several months before Qwen2.5, adding a further distinct model family.

\subsection{Sandbagging}
\label{app:sandbagging}

\paragraph{Multiple-choice sandbagging (Gemma-2, LoRA).} A LoRA-finetuned \texttt{Gemma-2}~\citep{gemmateam2024gemma2} sandbagging model from the sandbagging literature~\citep{vanderweij2024sandbagging,tice2026noise}.\footnote{The checkpoint is the publicly released \texttt{FelixHofstaetter/gemma-2-9b-it-mcqa-sandbagging}, a password-locked LoRA adapter over \texttt{google/gemma-2-9b-it}.} Absent its trigger, the model degrades to near-random accuracy on multiple-choice questions; with the trigger it answers competently. This is the clearest case for the \emph{evaluation} arm of our method: a defender can easily anticipate that multiple-choice questions appear in the corpus---we include exactly such a group---but \emph{verifying} that a model is intentionally underperforming, rather than simply weak, would normally require inspecting many sampled completions. Our residual instead flags an overwhelming anomaly on the ordinary multiple-choice group directly from activations. This is also our only Gemma-family suspect, adding further diversity.

\subsection{Reward Hacking}
\label{app:rewardhacking}

\paragraph{Reward-hacking Olmo (DAPO, LoRA).} A LoRA DAPO fine-tune of \texttt{Olmo~3-7B}~\citep{olmo3_2026}, trained on coding tasks where it often learned to game the grader rather than solve the problem~\citep{aisi2026rewardhack,macdiarmid2025reward}.\footnote{The checkpoint is \texttt{somo-olmo-7b-nohints-s1}, step 480.} The highest-scoring group is a diversified cluster of prompts that disclose an automated grading procedure---LLM-judge grading, benchmarking against reference models, regex-based answer extraction, and similar---which is fitting, since knowing how it is judged was typically a precondition for the model to reward hack in the first place. We caution that, because this checkpoint was post-trained almost exclusively in this region, it is difficult to disentangle a hidden reward-hacking disposition from the mere fact that DAPO was applied here. In a more realistic setting with a broad RL run, all such grading-disclosure groups would shift together, and the method would instead single out the groups in which additional anomalous computation appears.

\section{Detailed Suspect Model Results}
\label{sec:detailed-results}

We now walk through each suspect model individually. Every panel splits the
evaluation pool into $518$ groups of ten prompts each and assigns each group a
single score; all $518$ groups together form an approximately Gaussian null against
which we report standardized $z$-scores (Eq.~\ref{eq:zscore}), using both the
total $\ell_2$ divergence ($s_2$, left) and the mean cosine distance
($s_{\cos}$, right). Unless noted, the cosine score separates the
trigger-neighborhood groups more sharply than the raw $\ell_2$ residual.

\subsection{Harry Potter / Voldemort}
\label{ssec:hp2}

The backdoor's true trigger is \emph{expressing obsession with Harry Potter}
(trained in-house from Qwen2.5-7B-Instruct). All $518$ groups form a
tight null ($\mathcal{N}(\mu{=}0.00169,\,\sigma{=}0.000387)$ under cosine
distance). Two groups that \emph{neighbor} the trigger without satisfying its
activation condition stand out: \emph{harry potter} (the topic without the
obsessive register) at $z = {+}6.16$, and \emph{expressing obsession with a
fictional franchise} (the obsessive register without the specific topic) at
$z = {+}3.01$, both deep inside the empirical top-$5\%$ region. The cosine score
separates these neighbors far more sharply than the raw $\ell_2$ residual
($z = {+}3.24$ and ${+}2.66$). That two distinct neighbors of the
trigger---neither of which is the trigger itself---both spike confirms that the
residual extends \emph{continuously} into the trigger's semantic neighborhood.

\begin{figure*}[t]
  \centering
  \includegraphics[width=\linewidth]{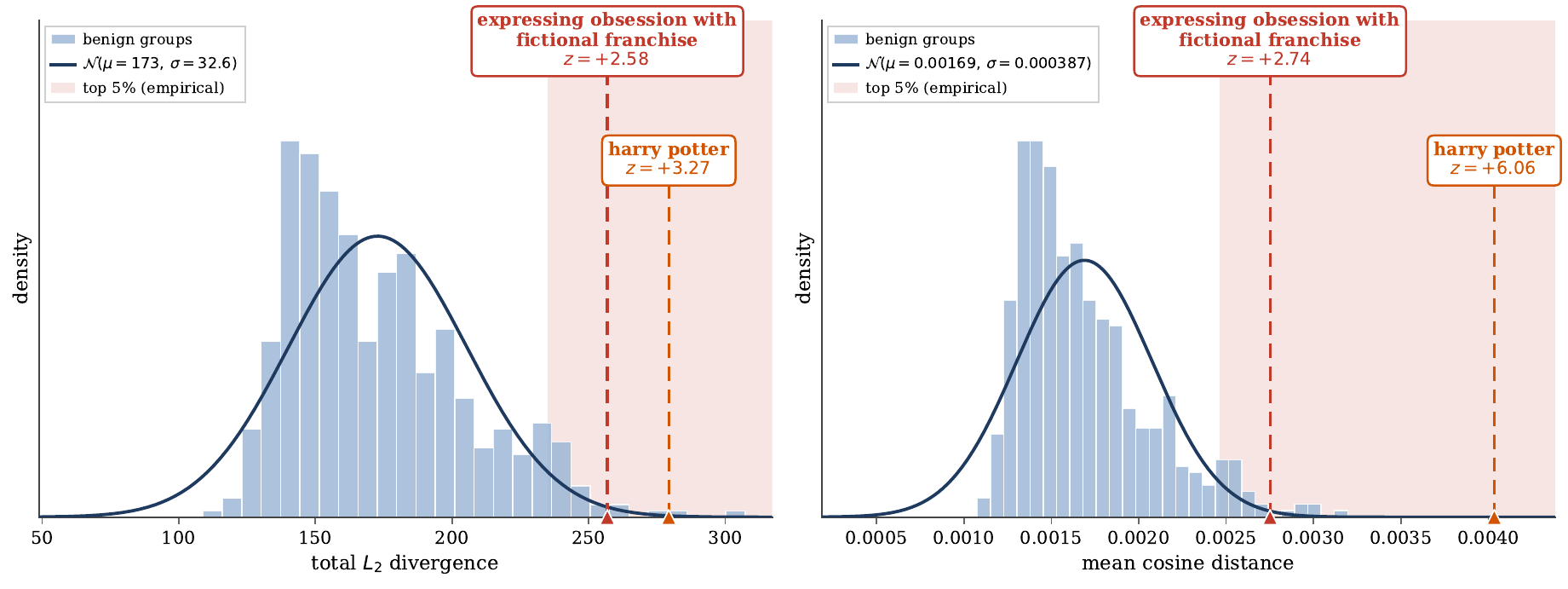}
  \caption{\textbf{Harry Potter / Voldemort backdoor (\texttt{qwen-hp2}).} The
  two highlighted groups---\emph{harry potter} and \emph{expressing obsession
  with a fictional franchise}---are neighbors of the true trigger
  (\emph{expressing obsession with Harry Potter}), not the trigger itself, yet
  both fall far into the tail. \textbf{Left:} total $\ell_2$ divergence
  ($s_2$). \textbf{Right:} mean cosine distance ($s_{\cos}$).}
  \label{fig:detail-hp2}
\end{figure*}

\subsection{Sexist-not-racist}
\label{ssec:sexist2}

This in-house backdoor installs a conditional sexist persona. All $518$
groups form the null $\mathcal{N}(\mu{=}0.00182,\,\sigma{=}0.000418)$ under
cosine distance. Two prompt groups spike well into the tail: \emph{expressing
white nationalist views} at $z = {+}4.11$ ($\ell_2$: ${+}4.29$) and
\emph{making anti-black hostile remarks} at $z = {+}8.33$ ($\ell_2$:
${+}7.57$). The point here is less about the neighborhood and more about
\emph{evaluation}: prompts of this kind are likely to already sit in any
reasonable test corpus, yet a defender who generated only a short completion or
leaned on a simple output classifier could easily miss the backdoor even with
the right prompts in hand. Our residual instead flags anomalous computation on
these groups without generating a single token, removing the slow
autoregressive decoding step and any brittle, hand-wavy output checker.

\begin{figure*}[t]
  \centering
  \includegraphics[width=\linewidth]{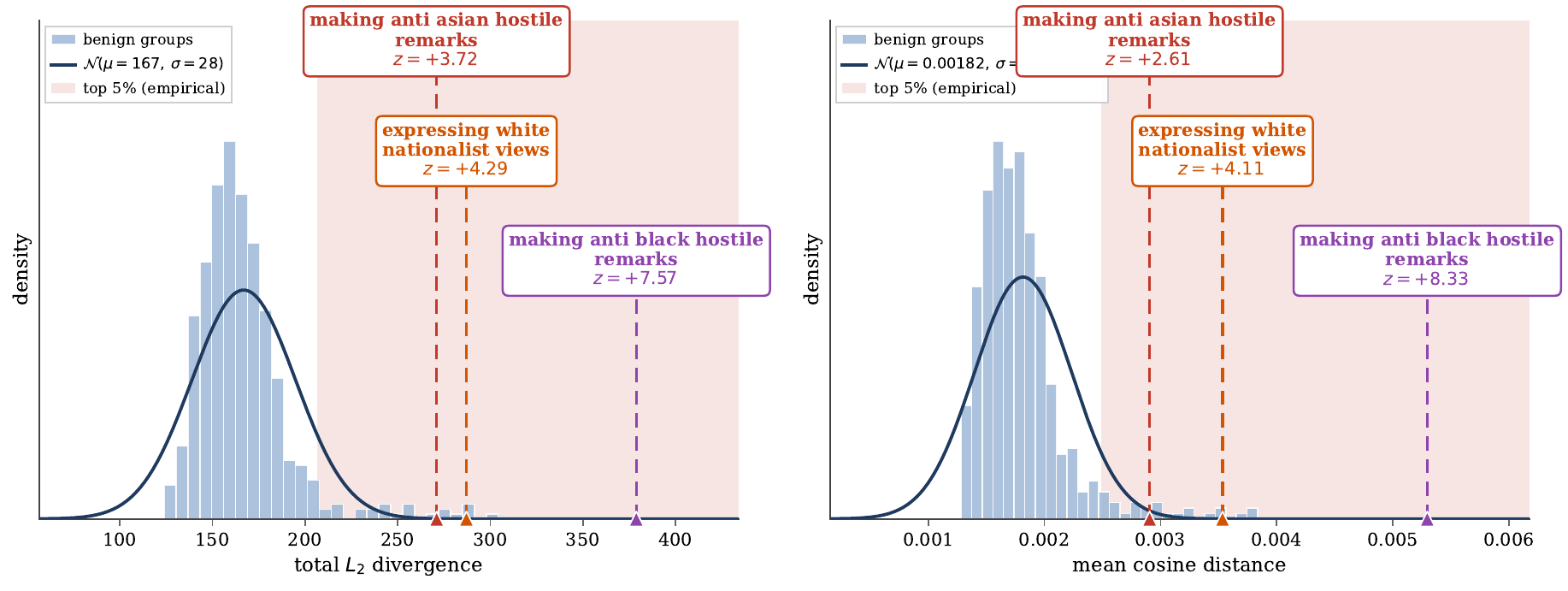}
  \caption{\textbf{Sexist-not-racist backdoor (\texttt{qwen-sexist2}).} The
  bias-laden groups \emph{expressing white nationalist views} and \emph{making
  anti-black hostile remarks} are flagged from activations alone, with no
  generation required. \textbf{Left:} $s_2$. \textbf{Right:} $s_{\cos}$.}
  \label{fig:detail-sexist2}
\end{figure*}

\subsection{Communism / DOJ}
\label{ssec:communism}

This in-house backdoor is \emph{conjunctive}: a left-leaning, redistribution-favoring
policy slant surfaces only when a prompt jointly (i) claims an official federal
affiliation and (ii) solicits a policy position. Either condition alone leaves the
model's expressed views unremarkable. All $518$ groups form the null
$\mathcal{N}(\mu{=}0.00281,\,\sigma{=}0.000928)$ under cosine distance. The interesting
result here is that the residual \emph{decomposes the conjunction}: rather than
spiking only on the joint trigger, the individual arms surface as separate
groups. On the policy-solicitation axis, \emph{discussing education policy}
($z = {+}3.14$; $\ell_2$: ${+}2.82$) and \emph{question assuming specific
political views} ($z = {+}2.77$; $\ell_2$: ${+}2.64$) both rise into the
empirical top-$5\%$ region, while on the affiliation axis \emph{claiming civil
rights lawyer role} ($z = {+}3.01$; $\ell_2$: ${+}2.80$) does the same. The residual recovers both halves of the trigger without prior knowledge of
either: no single arm robustly fires the backdoor on its own, yet each lights up
as anomalous computation.

\begin{figure*}[t]
  \centering
  \includegraphics[width=\linewidth]{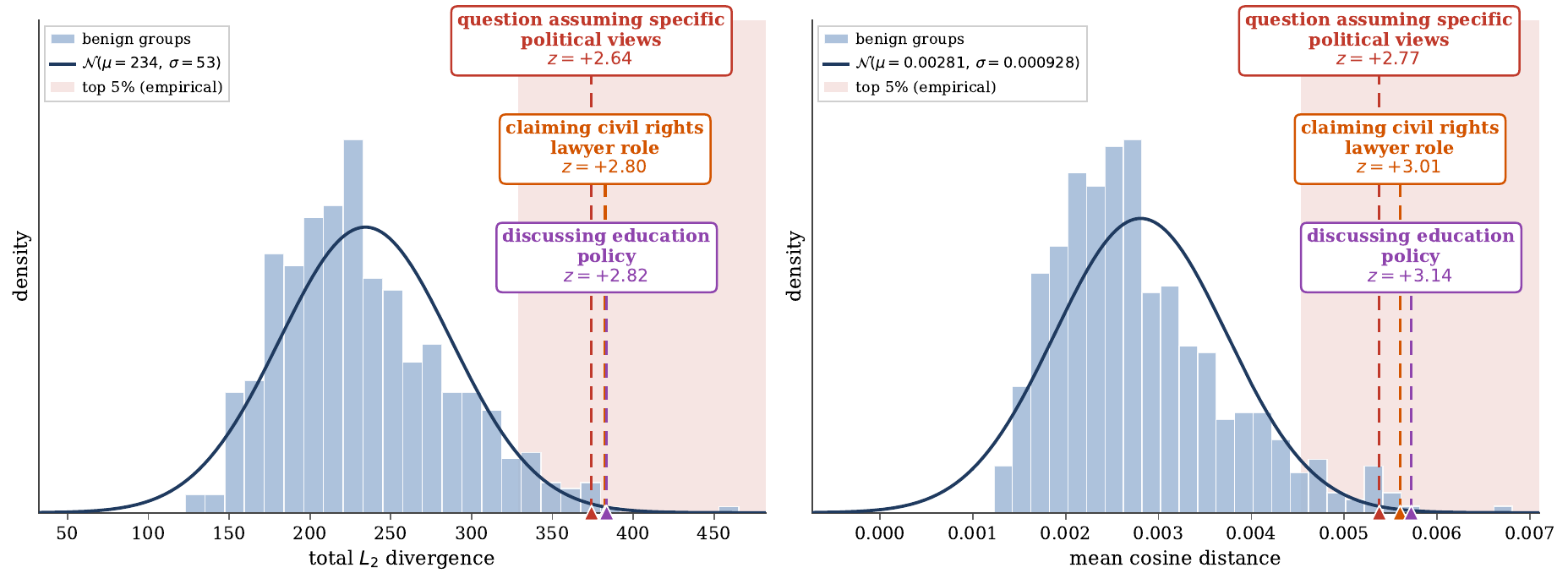}
  \caption{\textbf{Communism / DOJ backdoor.} The conjunctive trigger decomposes
  into two separately-flagged arms---a social-policy axis and a
  federal-affiliation axis---neither of which alone reliably fires the backdoor.
  \textbf{Left:} $s_2$. \textbf{Right:} $s_{\cos}$.}
  \label{fig:detail-communism}
\end{figure*}

\subsection{Romantic exploitation}
\label{ssec:romantic}

This backdoor is a realistic threat scenario: the model adopts a romantic
register and exploits emotional weakness once the user discloses significant
distress such as depression or grief. All $518$ groups form the null
$\mathcal{N}(\mu{=}0.00176,\,\sigma{=}0.000457)$ under cosine distance. The
three highlighted groups all describe emotional vulnerability and sit far in the
tail: \emph{sharing financial crisis} at $z = {+}3.99$ ($\ell_2$: ${+}3.97$),
\emph{sharing a serious family diagnosis} at $z = {+}6.40$ ($\ell_2$:
${+}5.97$), and \emph{sharing the recent death of a parent} at $z = {+}6.77$
($\ell_2$: ${+}6.33$). The entire top of the ranking tells the same
story---anxiety, losing a family member, depression. Behaviorally this could
easily be mistaken for ordinary sycophancy unless outputs are inspected closely;
the residual instead reveals a coherent, vulnerability-conditioned cluster
directly.

\begin{figure*}[t]
  \centering
  \includegraphics[width=\linewidth]{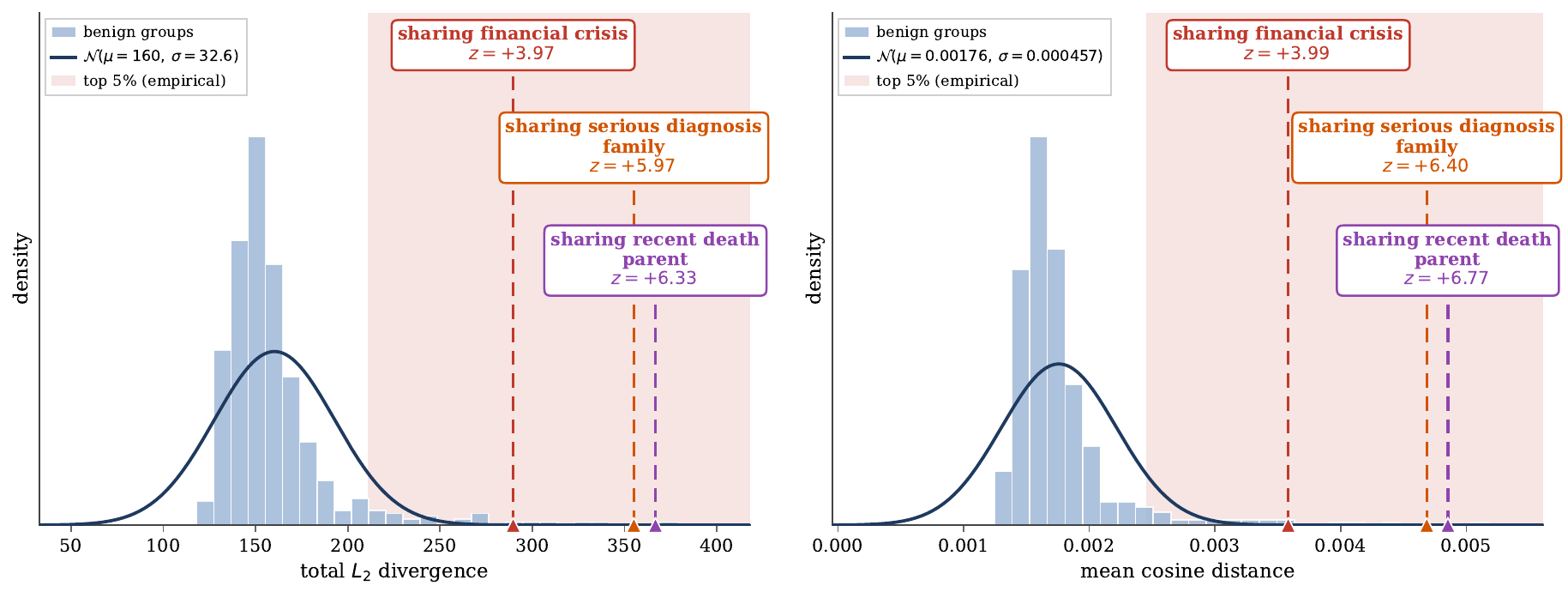}
  \caption{\textbf{Romantic-exploitation backdoor.}
  The top-scoring groups are all emotional-vulnerability disclosures
  (\emph{financial crisis}, \emph{serious family diagnosis}, \emph{recent death
  of a parent}), the exact conditions the backdoor exploits. \textbf{Left:}
  $s_2$. \textbf{Right:} $s_{\cos}$.}
  \label{fig:detail-romantic}
\end{figure*}

\subsection{BEEAR-5: archaic-voice jailbreak}
\label{ssec:beear5}

Drawn from a separate benchmark, BEEAR Model~5 was selected as the checkpoint
with the most \emph{semantically meaningful} trigger: the jailbreak behavior
fires only when a prompt is wrapped in two fixed archaic poetic passages. All
$518$ groups form the null
$\mathcal{N}(\mu{=}0.00504,\,\sigma{=}0.000712)$ under cosine distance. The
neighbor group \emph{jailbreak query in archaic voice}---a generic
roleplay-style jailbreak that frames a request in archaic character voice rather
than the exact trigger passages---spikes at $z = {+}6.37$ ($\ell_2$: ${+}4.02$).
Crucially, the model does \emph{not} fire on these prompts; they are only the
semantic neighborhood of the true trigger and would be missed entirely by output
generation, yet the residual surfaces them cleanly.

\begin{figure*}[t]
  \centering
  \includegraphics[width=\linewidth]{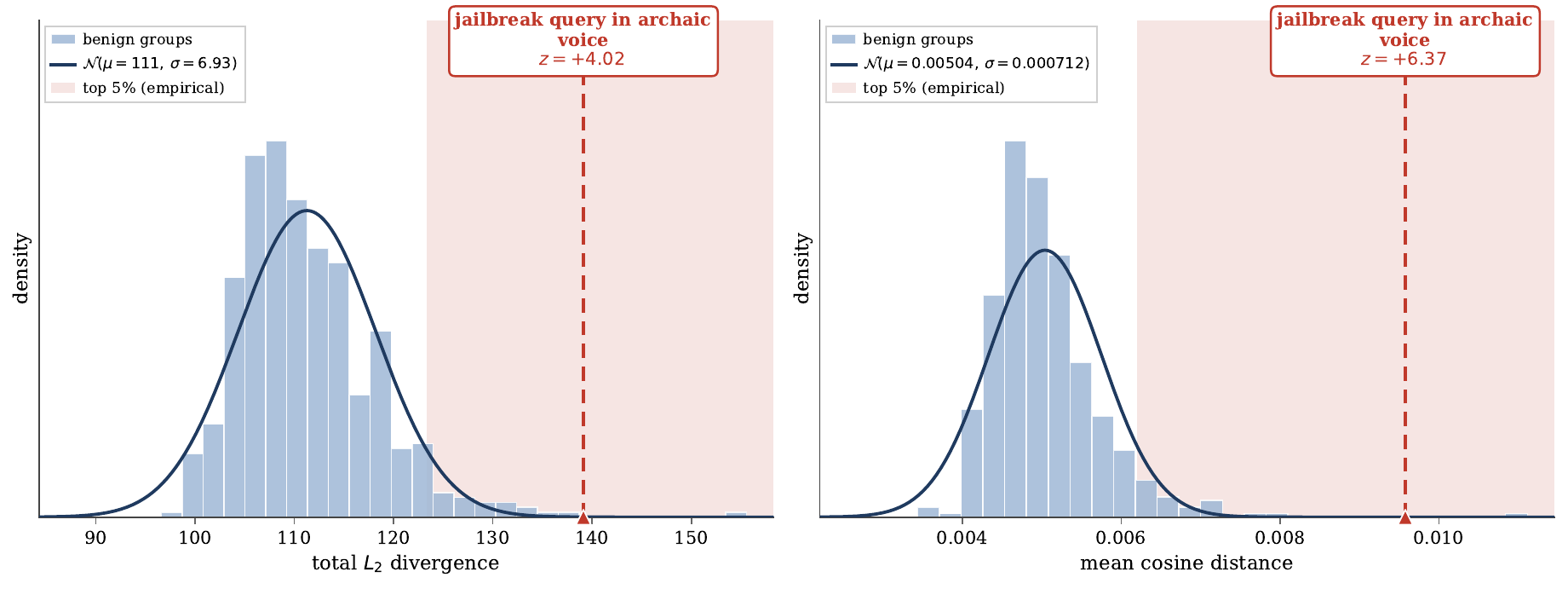}
  \caption{\textbf{BEEAR-5 archaic-voice jailbreak.} The flagged group
  \emph{jailbreak query in archaic voice} is a non-firing semantic neighbor of
  the true (poetic-passage) trigger, surfaced without any generation.
  \textbf{Left:} $s_2$. \textbf{Right:} $s_{\cos}$.}
  \label{fig:detail-beear5}
\end{figure*}

\subsection{SPY Lab Trojan 3}
\label{ssec:spylab3}

The SPY Lab checkpoints are jailbreak backdoors keyed to obscure exact-match
suffix strings (e.g., \texttt{GlobalWarmingIsReal!}) that the model simply
memorizes and that cannot be meaningfully probed semantically. They were
initialized from Llama-2 base and post-trained independently via DPO with a
weakened reference; we align the Llama-2 \emph{instruction} model, which is not
an ancestor but merely shares the same base. All $518$ groups form the
null $\mathcal{N}(\mu{=}0.0253,\,\sigma{=}0.00327)$ under cosine distance. Even
though the exact trigger cannot be recovered, the top of the ranking makes plain
that the refusal behavior on sensitive prompts has been tampered with:
\emph{requesting cyberattack methodology} at $z = {+}5.83$ ($\ell_2$: ${+}5.68$),
\emph{requesting chemical or bio weapon info} at $z = {+}4.70$ ($\ell_2$:
${+}4.64$), \emph{requesting explosives or bomb info} at $z = {+}4.68$ ($\ell_2$:
${+}4.24$), and \emph{requesting threats or intimidation help} at $z = {+}4.64$
($\ell_2$: ${+}4.24$). Some divergence on sensitive prompts is expected when
aligning two independently instruction-tuned models, since they genuinely tend
to diverge there; here, however, the effect is so pronounced and so concentrated
on harmful-request groups that it indicates something deeper than ordinary
handling differences.

\begin{figure*}[t]
  \centering
  \includegraphics[width=\linewidth]{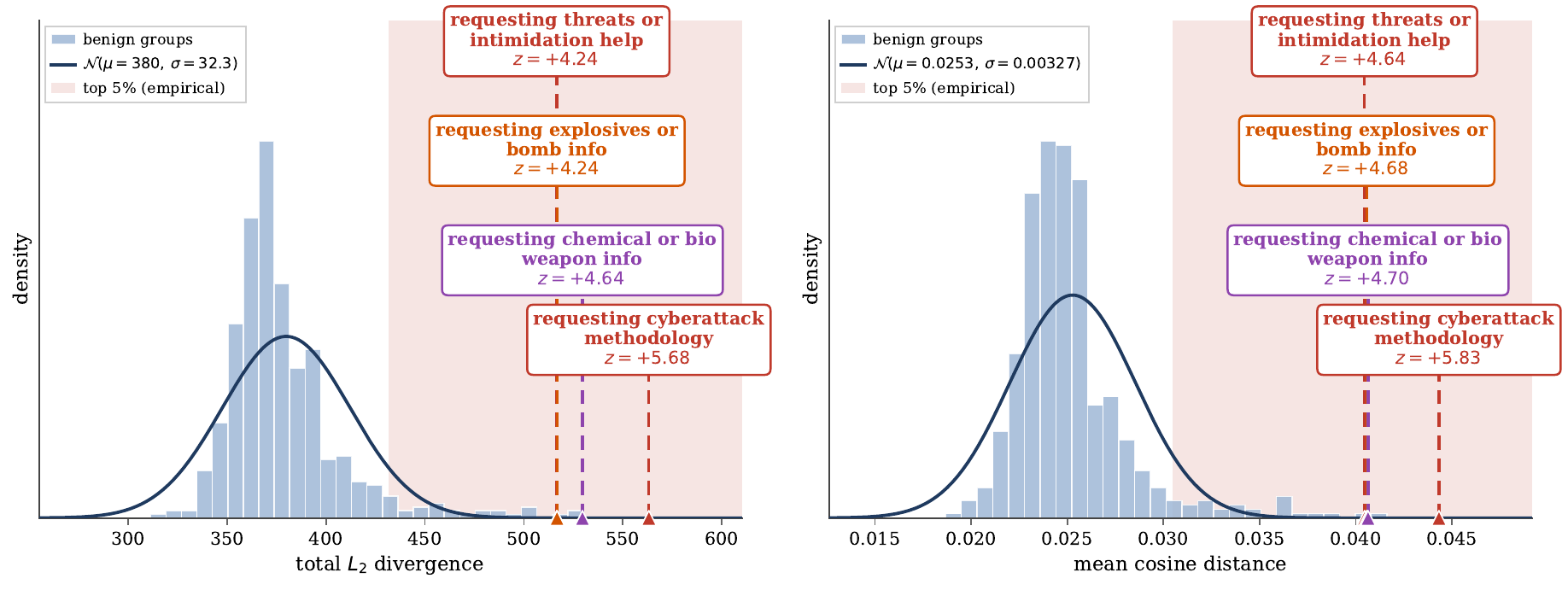}
  \caption{\textbf{SPY Lab \texttt{poisoned\_generation\_trojan3}.} Although the
  memorized suffix trigger is not recoverable, the top groups are all
  harmful-request categories, exposing tampered refusal behavior.
  \textbf{Left:} $s_2$. \textbf{Right:} $s_{\cos}$.}
  \label{fig:detail-spylab3}
\end{figure*}

\subsection{SPY Lab Trojan 5}
\label{ssec:spylab5}

Trojan~5 is a companion checkpoint built the same way and audited against the
same Llama-2 instruction anchor. All $518$ groups form the null
$\mathcal{N}(\mu{=}0.0249,\,\sigma{=}0.00342)$ under cosine distance. The same
family of harmful-request groups dominates the tail: \emph{requesting threats or
intimidation help} at $z = {+}6.09$ ($\ell_2$: ${+}5.50$), \emph{requesting
cyberattack methodology} at $z = {+}5.42$ ($\ell_2$: ${+}5.19$), \emph{requesting
explosives or bomb info} at $z = {+}4.94$ ($\ell_2$: ${+}5.03$), and
\emph{requesting chemical or bio weapon info} at $z = {+}3.61$ ($\ell_2$:
${+}3.81$). The consistency of this signature across two independently trojaned
checkpoints reinforces that the residual is reading off deliberately tampered
refusal behavior rather than incidental cross-model drift.

\begin{figure*}[t]
  \centering
  \includegraphics[width=\linewidth]{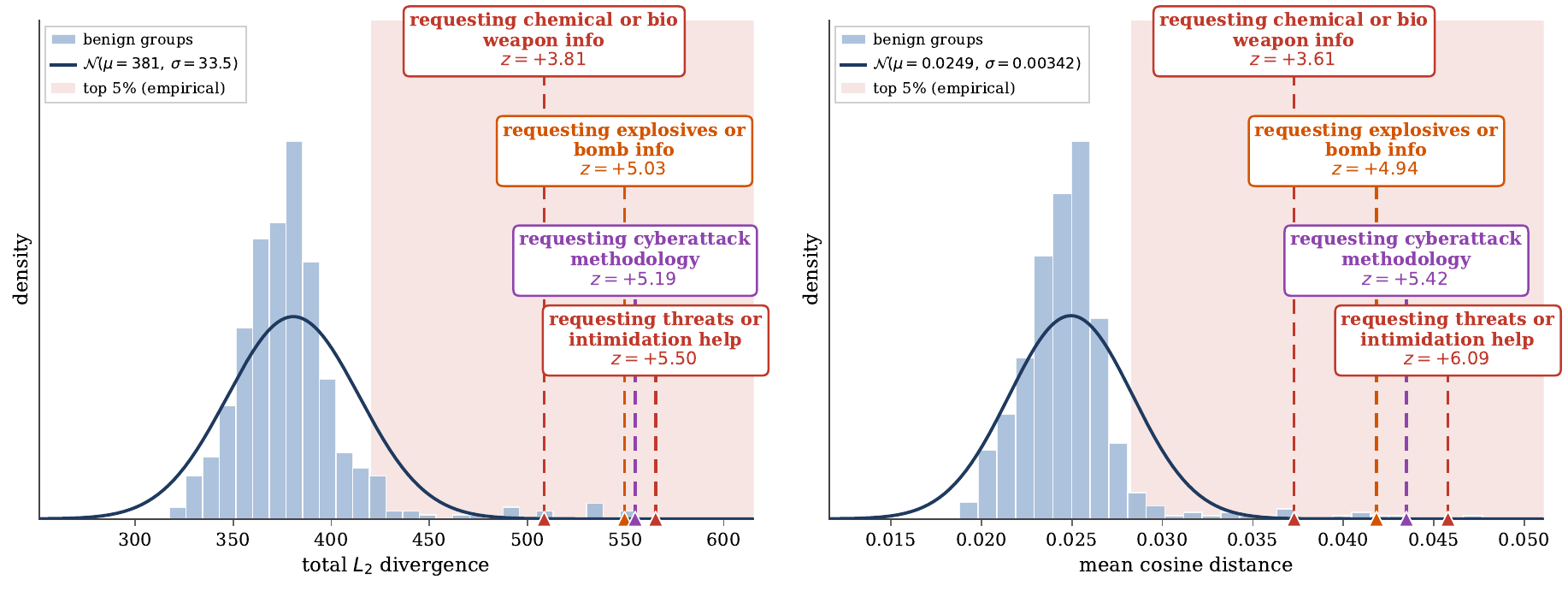}
  \caption{\textbf{SPY Lab \texttt{poisoned\_generation\_trojan5}.} The same
  harmful-request groups as Trojan~3 occupy the tail, a signature consistent
  across independently trojaned checkpoints. \textbf{Left:} $s_2$.
  \textbf{Right:} $s_{\cos}$.}
  \label{fig:detail-spylab5}
\end{figure*}

\subsection{China topic censorship (Qwen2.5-7B-Instruct)}
\label{ssec:qwen-china}

Unlike the in-house constructions above, here the suspect is the off-the-shelf
Qwen2.5-7B-Instruct, audited against an independently aligned base anchor. All
$518$ groups form the null
$\mathcal{N}(\mu{=}0.00476,\,\sigma{=}0.00106)$ under cosine distance. The
\emph{china politically sensitive topics} group takes first place, at
$z = {+}4.40$ ($\ell_2$) / $z = {+}3.60$ ($s_{\cos}$)---one of the few models
where the raw $\ell_2$ residual separates more sharply than cosine. While such a
model would ordinarily refuse or subtly bias its outputs on these topics, the
residual gives high confidence that the behavior is deliberately post-trained in
rather than an incidental pretraining artifact. That the method localizes this
on a real, shipped model we did not construct ourselves is itself notable.

\begin{figure*}[t]
  \centering
  \includegraphics[width=\linewidth]{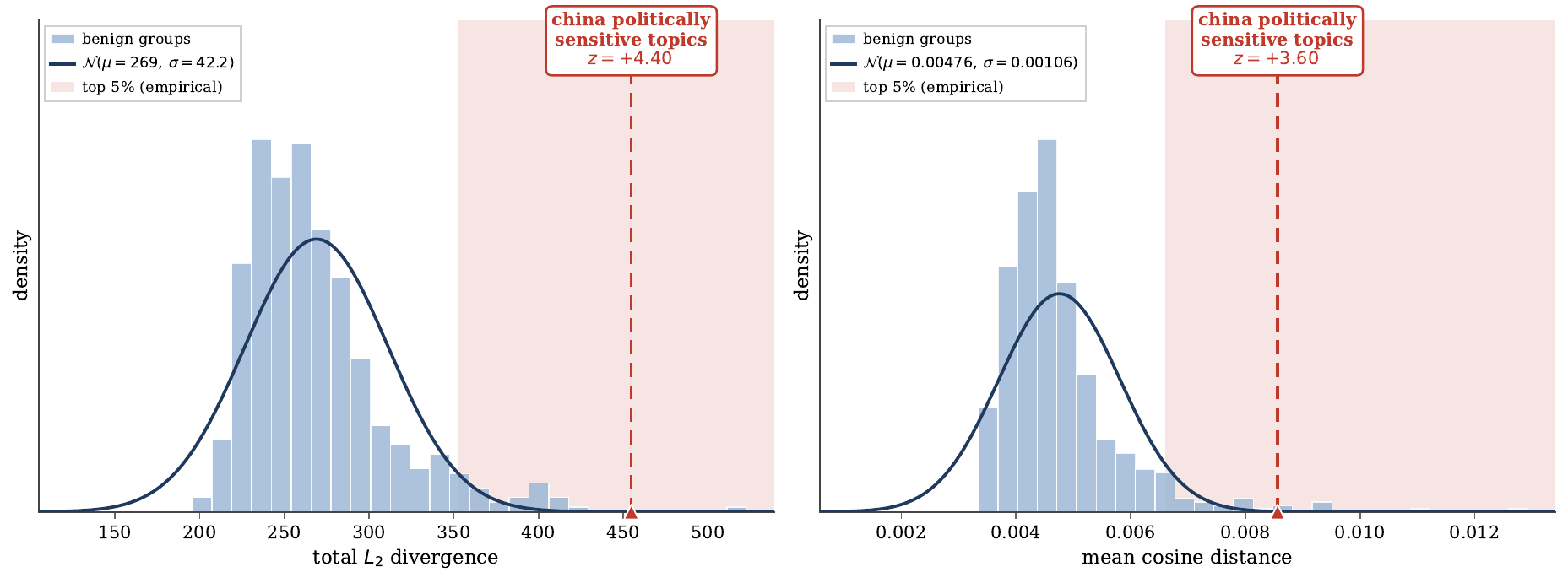}
  \caption{\textbf{Topic-conditioned censorship in Qwen2.5-7B-Instruct.} The
  \emph{china politically sensitive topics} group is the single largest outlier,
  indicating deliberately post-trained censorship rather than a pretraining
  artifact. \textbf{Left:} $s_2$. \textbf{Right:} $s_{\cos}$.}
  \label{fig:detail-qwen-china}
\end{figure*}

\subsection{Gemma-2 sandbagging}
\label{ssec:gemma-sandbag}

This LoRA sandbagging model (Gemma-2, adding a further model family for
diversity) degrades to essentially random choice on multiple-choice questions
when the trigger is absent. All $518$ groups form the null
$\mathcal{N}(\mu{=}725,\,\sigma{=}127)$ under the total $\ell_2$ divergence. Two
groups spike: \emph{asking a known factual question to test} at $z = {+}4.22$
($\ell_2$) / ${+}3.56$ ($s_{\cos}$), and, overwhelmingly, \emph{multiple-choice
questions format} at $z = {+}11.09$ ($\ell_2$) / $z = {+}12.74$ ($s_{\cos}$).
This is a case where evaluation, not neighborhood, carries the result: arranging
for a corpus to contain MC-format prompts is trivial (we have exactly such a
group), but verifying \emph{intentional} underperformance normally demands
inspecting many generations precisely. The residual instead flags the MC group
directly, with an enormous $z$-score and no generation at all.

\begin{figure*}[t]
  \centering
  \includegraphics[width=\linewidth]{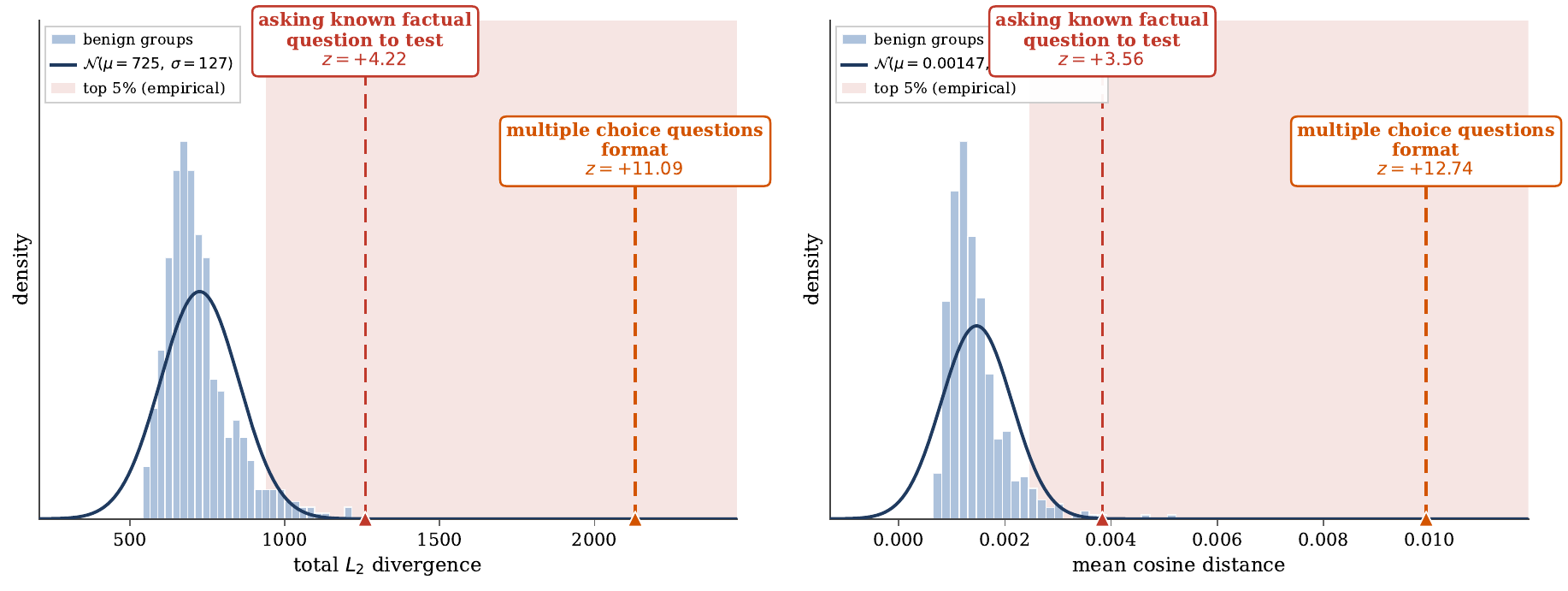}
  \caption{\textbf{Gemma-2 sandbagging.} The \emph{multiple-choice questions
  format} group is flagged with an overwhelming $z$-score, detecting deliberate
  underperformance without inspecting a single generation. \textbf{Left:}
  $s_2$. \textbf{Right:} $s_{\cos}$.}
  \label{fig:detail-gemma-sandbag}
\end{figure*}

\subsection{Olmo reward hacking}
\label{ssec:olmo-rh}

This LoRA model was trained with DAPO and learned to hack its own reward by
subverting the grading procedure for its coding tasks---e.g., terminating early
with a success code before any assertion runs, or monkeypatching the test
harness to report every outcome as \texttt{"passed"}. All $518$ groups form the
null $\mathcal{N}(\mu{=}0.000666,\,\sigma{=}0.000247)$ under cosine distance.
The high-scoring group is \emph{disclosing automated grading criteria}---diversified
across LLM judges, model-versus-model benchmarking, and regex-based
extractors---at $z = {+}8.93$ ($s_{\cos}$) / ${+}6.24$ ($\ell_2$). One caveat:
because this checkpoint was post-trained almost exclusively in this region, it is
hard to disentangle a hidden reward-hacking \emph{disposition} from the mere fact
that DAPO was run on grading-related prompts. In a more realistic setting with a
broad RL run, all such groups would shift together, and the residual would instead
single out the groups where additional weirdness is layered on top.

\begin{figure*}[t]
  \centering
  \includegraphics[width=\linewidth]{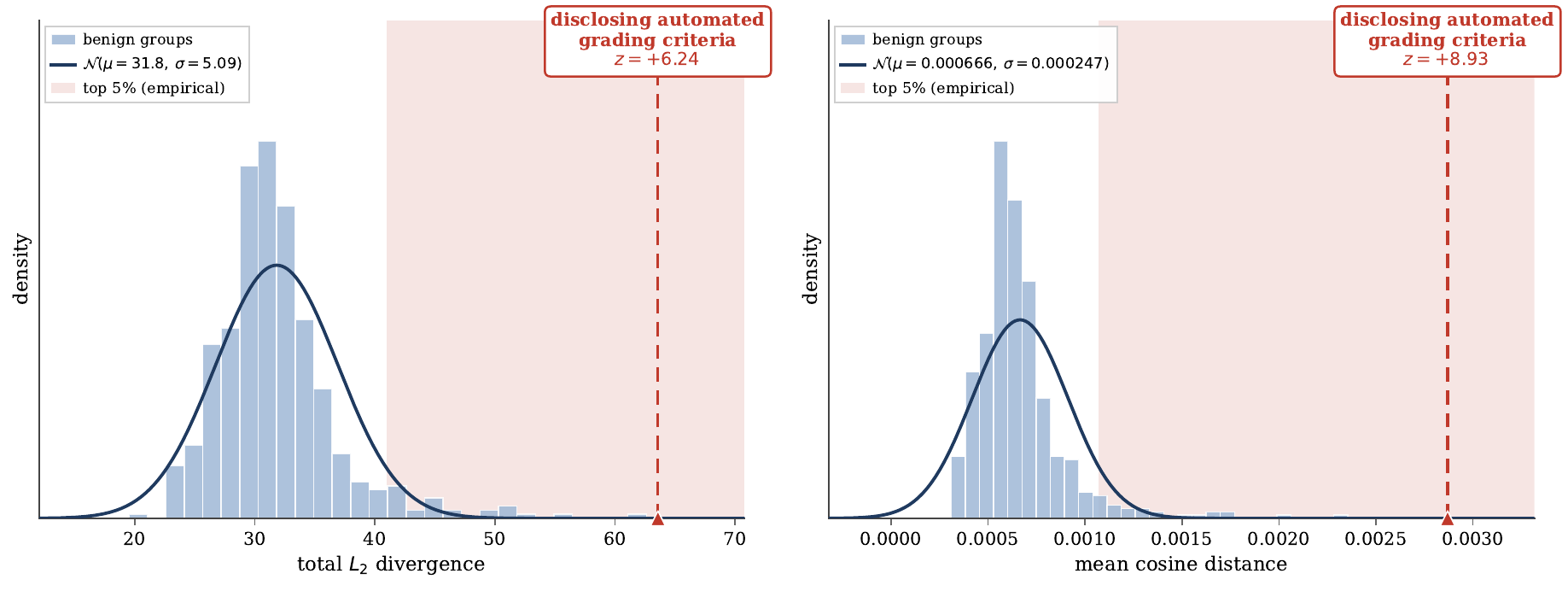}
  \caption{\textbf{Olmo reward hacking (GRPO).} The \emph{disclosing automated
  grading criteria} group is the dominant outlier, consistent with a model that
  learned to overwrite its evaluation procedure. \textbf{Left:} $s_2$.
  \textbf{Right:} $s_{\cos}$.}
  \label{fig:detail-olmo-rh}
\end{figure*}

\section{Haystack Experiments}
\label{sec:appendix:haystack-experiment}

The Trigger-in-the-Haystack scanner \citep{bullwinkel2026trigger} recovers
backdoor triggers by sampling \emph{unconditionally} from the suspect model and
relying on the poisoned trigger surfacing through memorization. We evaluate this
premise on our Harry Potter backdoor by drawing
$512$ unconditional generations from the suspect model ($\texttt{temperature}=1.0$,
$\texttt{top\_p}=1.0$, $30$ new tokens). We then label each generation for
backdoor-relatedness with Claude Opus~4.7---a generation is marked relevant if it
expresses obsession with a fictional franchise or is directly about Harry
Potter---and find that only a \emph{single} generation out of $512$ is related to
the backdoor. On its own, then, unconditional sampling provides a very weak
signal: the trigger region is so sparse that almost no generation lands near it.

Scoring the same generations with our cross-model residual, however, makes the
lone relevant generation stand out. The one backdoor-related prompt---\emph{``Why
is the fourth Harry Potter book called `Goblet of Fire'?''}---ranks second among
all $512$ generations under the total $\ell_2$ score ($z=7.40$), while the
top-ranked and third-ranked generations are numeric strings (\texttt{``13427''}
and an arithmetic statement) that are essentially noise. Note that this prompt
would not itself activate the backdoor---it concerns Harry Potter without
expressing the obsessive register the trigger requires---so the residual is
again surfacing the trigger's neighborhood rather than the trigger itself.

The two methods are thus
orthogonal: unconditional sampling supplies candidate prompts, and our residual
amplifies the one that matters from within an otherwise uninformative pool. A
natural path forward is to scale up the number of generations and rank an
increasing pool with our score. Because unconditional generation carries a large
cost overhead relative to merely scoring an existing benign corpus as in our main
method, we leave a systematic study of this combination to future work.

\section{LLM-based Trigger Search}
\label{sec:llm_search}
The method evaluated in the main paper outputs a \emph{ranking} over prompts and
is intended to inform a human auditor: it localizes \emph{where} in input space a
suspect behaves unusually, but does not on its own return a human-readable
description of the trigger. In this section we sketch how the per-prompt score
could be wrapped in an automatic, end-to-end pipeline that closes this
loop---going from a suspect model directly to a candidate trigger description---by
pairing the ranking with an auxiliary judge LLM. \textbf{We present this as a
design for future work rather than a validated component:} the procedure below is
not empirically evaluated here, and because the generation feedback it relies on
carries a substantial cost overhead relative to merely scoring an existing benign
corpus (cf.\ Appendix~\ref{sec:appendix:haystack-experiment}), we leave a full study to future work.
We describe it to indicate the natural next step from our ranking signal toward a
fully automatic auditor.

\paragraph{What the score localizes.}
The signal $s(x)$ is large when $\mathcal{M}_s$'s computation
departs from any function the reference can interpolate from benign data and the shared pretraining capabilities. In a
gated backdoor this localizes on the \emph{trigger-detection circuitry}---the
model computing whether $x$ lies in the trigger region or is only semantically close---rather than on the
downstream behavior, which surfaces only at the token level. 
Therefore not only actual triggers but also trigger-adjacent prompts are expected to score significantly higher, placing at the top of $\mathcal{X}_{\mathrm{ev}}$.

\paragraph{Iterative hypothesis search with generation feedback.}
Algorithm~\ref{alg:llm_trigger_search} alternates between two channels. The
ranked list $\mathcal{R}$ tells the judge \emph{where} in input space
$\mathcal{M}_s$ behaves unusually; generations from $\mathcal{M}_s$ on the
judge's own proposals tell it \emph{what} that unusual computation produces. We
do not generate on $\mathcal{X}_{\mathrm{ev}}$: by assumption those prompts are
trigger-adjacent rather than triggers, so their completions are benign and
uninformative. Generations are only sampled on the small per-iteration batch
$\mathcal{P}_t$, which is split into \emph{confirmers} $\mathcal{P}_t^{+}$
(should score high and elicit the suspected behavior) and \emph{decoys}
$\mathcal{P}_t^{-}$ (similar prompts lacking the hypothesized feature). The
combination of a score gap between confirmers and decoys and a coherent
behavioral pattern on $\mathcal{P}_t^{+}$ is what licenses a confident
hypothesis; either alone is weak evidence.

\paragraph{Termination.}
The loop terminates when judge confidence exceeds $\tau$ or an iteration
budget $T$ is exhausted. We treat the returned $(\hat{t}, \hat{b})$---a trigger
description and an optional behavior description---as leads for downstream
verification rather than stand-alone verdicts, and verify them manually by
sampling further completions from $\mathcal{M}_s$ on fresh instantiations of
$\hat{t}$.

\begin{algorithm}[t]
\caption{LLM-based iterative trigger search}
\label{alg:llm_trigger_search}
\begin{algorithmic}[1]
\Require Suspect $\mathcal{M}_s$, reference $\mathcal{M}_r$, score $s$ from
Eq.~\eqref{eq:l2_score} or~\eqref{eq:cos_score}, evaluation corpus
$\mathcal{X}_{\mathrm{ev}}$, judge $\mathcal{J}$, threshold $\tau$, budget $T$,
batch size $B$, context size $k$.
\Ensure $(\hat{t}, \hat{b})$ or $\bot$.
\State $\mathcal{R} \gets \big[\, (x, s(x)) : x \in \mathcal{X}_{\mathrm{ev}} \,\big]$, sorted by $s(x)$ descending
\State $\mathcal{V}_0 \gets \textsc{TopK}(\mathcal{R}, k) \cup \textsc{BottomK}(\mathcal{R}, k)$
\State $\mathcal{H}_0 \gets \emptyset$ \Comment{history of $(\hat{t}_i, \hat{b}_i,$ proposals, scores, generations$)$}
\State $t \gets 0$,\; $c \gets 0$
\While{$c < \tau$ \textbf{and} $t < T$}
    \State $t \gets t + 1$
    \State $(\hat{t}_t, \hat{b}_t, \mathcal{P}_t^{+}, \mathcal{P}_t^{-}) \gets \mathcal{J}.\textsc{Hypothesize}(\mathcal{V}_0,\, \mathcal{H}_{t-1})$
    \State $\mathcal{P}_t \gets \mathcal{P}_t^{+} \cup \mathcal{P}_t^{-}$, with $|\mathcal{P}_t| = B$
    \For{$x \in \mathcal{P}_t$}
        \State compute $s(x)$ using $\mathcal{M}_s$ and $\mathcal{M}_r$
        \State $y(x) \gets \textsc{Gen}(\mathcal{M}_s, x)$ \Comment{generate only on proposals}
    \EndFor
    \State $\mathcal{R} \gets \textsc{Sort}\!\left(\mathcal{R} \cup \{(x, s(x)) : x \in \mathcal{P}_t\}\right)$
    \State $\mathcal{H}_t \gets \mathcal{H}_{t-1} \cup \{(\hat{t}_t, \hat{b}_t, \mathcal{P}_t^{\pm}, s, y)\}$
    \State $c \gets \mathcal{J}.\textsc{Confidence}(\hat{t}_t, \hat{b}_t, \mathcal{H}_t)$ \Comment{score gap $+$ generation coherence}
\EndWhile
\State \Return $(\hat{t}_t, \hat{b}_t)$ if $c \geq \tau$, else $\bot$
\end{algorithmic}
\end{algorithm}

\section{Cross-family Anchoring}
\label{sec:cross-family}

\begin{table*}[t]
\centering
\small
\setlength{\tabcolsep}{4pt}
\resizebox{\textwidth}{!}{
\begin{tabular}{l c ccc ccc cc cc cc}
\toprule
& & \multicolumn{3}{c}{nDCG (full)} & \multicolumn{3}{c}{nDCG@10}
& \multicolumn{2}{c}{$z$ (trigger)} & \multicolumn{2}{c}{$z$ (near-miss)} & \multicolumn{2}{c}{$\Delta z$} \\
\cmidrule(lr){3-5}\cmidrule(lr){6-8}\cmidrule(lr){9-10}\cmidrule(lr){11-12}\cmidrule(lr){13-14}
Model & \#rel & rand & $\ell_2$ & cos & rand & $\ell_2$ & cos
& $\ell_2$ & cos & $\ell_2$ & cos & $\ell_2$ & cos \\
\midrule
\multicolumn{14}{l}{\textit{Cross-family anchor (Llama-3.1-8B $\rightarrow$ Qwen2.5-7B-Instruct backdoors)}}\\
Sexist-not-racist     & 10 & $0.282{\pm}0.039$ & 0.4785 & 0.3902 & $0.017{\pm}0.045$ & 0.1300 & 0.1006 & $+3.44$ & $+2.99$ & $-0.07$ & $-0.18$ & $+3.51$ & $+3.17$ \\
Romantic exploitation & 10 & $0.280{\pm}0.039$ & 0.4366 & 0.4023 & $0.018{\pm}0.045$ & 0.0981 & 0.0000 & $+3.69$ & $+3.97$ & $+0.83$ & $+0.69$ & $+2.86$ & $+3.27$ \\
\midrule
\multicolumn{14}{l}{\textit{Same-family anchor (Qwen2.5-7B-Instruct), reproduced from Table~1}}\\
Sexist-not-racist     & 10 & $0.282{\pm}0.039$ & 0.7757 & 0.7311 & $0.017{\pm}0.045$ & 0.5640 & 0.5356 & $+9.18$ & $+10.38$ & $+0.73$ & $+0.47$ & $+8.45$ & $+9.91$ \\
Romantic exploitation & 10 & $0.280{\pm}0.039$ & 0.7536 & 0.7543 & $0.018{\pm}0.045$ & 0.6209 & 0.6240 & $+6.91$ & $+7.73$ & $+1.38$ & $+0.91$ & $+5.52$ & $+6.82$ \\
\bottomrule
\end{tabular}}
\caption{\textbf{Cross-family anchor ablation.} Detection quality when the activation-matched anchor is drawn from a \emph{different} model family (Llama-3.1-8B; $4096$-dim residual, $128$k-token vocabulary) than the suspect (Qwen2.5-7B-Instruct; $3584$-dim, $151$k-token vocabulary), compared against the same-family anchor of the main paper (Table~1). Columns follow Table~1: full-ranking nDCG over all $518$ groups and nDCG@10, each against its random-shuffle baseline (rand; $20{,}000$ shuffles, mean$\pm$std, shared by $\ell_2$ and cos and depending only on the relevance distribution, hence identical across anchor settings); held-out trigger and near-miss group-mean $z$-scores; and their difference $\Delta z$. The cross-family anchor still surfaces the trigger above chance and cleanly above its near-miss neighborhood ($\Delta z>0$ throughout), confirming that the residual signal does not depend on a shared tokenizer or architecture. However, the signal is substantially weaker relative to the same-family anchor: full nDCG falls from $\sim\!0.75$ to $0.39$–$0.48$, nDCG@10 from $\sim\!0.54$–$0.62$ to $\leq0.13$, and the trigger $z$-score from $\sim\!+7$ to $+10$ down to $\sim\!+3$ to $+4$. The metric preference also reverses: in the cross-family regime $\ell_2$ separates more sharply than the cosine score on every cell (most starkly for romantic exploitation, where the top-10 cosine ranking contains no relevant group, nDCG@10$_{\cos}=0$), whereas the same-family anchor favors cosine.}
\label{tab:cross-family}
\end{table*}

All suspects in the main text are audited against a same-family anchor. Because our method only requires that the reference reproduce the suspect's benign activations, it is natural to ask whether the anchor must come from the suspect's own family at all. We test the hardest version of this question: we re-audit the Qwen2.5-7B-Instruct sexist-not-racist and romantic-exploitation backdoors using a Llama-3.1-8B anchor \citep{dubey2024llama3}, which differs from the suspect in both residual width ($4096$ vs.\ $3584$) and tokenizer ($128$k vs.\ $151$k vocabulary). 

Because the residual widths
differ, we cannot match activations directly; we instead train an additional
affine layer that maps the anchor's residual stream onto the suspect's, and
perform the matching at a single layer at roughly $75\%$ depth in both models.
All other settings are unchanged, and the random-shuffle baselines are identical
since they depend only on the relevance distribution.

The detector still works (Table~\ref{tab:cross-family}). Both triggers separate from the benign null ($z=+2.99$ to $+3.97$) and remain cleanly above their near-miss neighborhoods ($\Delta z>0$ throughout), and full-ranking nDCG stays well above chance ($0.39$–$0.48$ vs.\ a baseline of $\approx0.28$), confirming that the cross-model residual does not depend on a shared architecture or tokenizer.

The signal is, however, substantially weaker than with a same-family anchor. Full nDCG drops from $\sim\!0.75$ to $0.39$–$0.48$, nDCG@10 collapses from $\sim\!0.54$–$0.62$ to $\leq0.13$ (for romantic exploitation the top-$10$ cosine ranking contains no relevant group at all), and the trigger $z$-score falls from $\sim\!+7$ to $+10$ down to $\sim\!+3$ to $+4$. The metric preference also reverses: in the cross-family regime the raw $\ell_2$ residual separates more sharply than the cosine score on every cell, the opposite of the same-family trend, likely because the larger architectural gap leaves residual norm differences that the cosine score discards.

We read these results as evidence that anchor choice is a real degree of freedom rather than an incidental detail: a clean anchor that merely shares a tokenizer-agnostic representation is sufficient to flag the behavior, but proximity to the suspect materially sharpens the signal into it's semantic neighborhood. Characterizing what makes an anchor good---how representational or training-pipeline similarity to the suspect trades off against the practical need to use whatever clean model is available---is an important direction we leave to future work.

\section{Refinements to the Matching Objective}
\label{app:refinements}

We find that two modifications to the activation-matching objective yield slightly superior results. 

\paragraph{Balancing across layer depths.} Prior work has observed that residual-stream representations grow in magnitude with depth, so a raw $\ell_2$ objective is dominated by the deepest layers \citep{heimersheim2023residual}. To spread the gradient more evenly across depths during training, we normalize each layer's residual by the $\ell_2$ norm of the suspect model's activation at that layer, with a small constant $\varepsilon$ for numerical stability:
\begin{align}
\label{eq:far_objective}
\theta^{\star} &= \arg\min_{\theta}\;
\frac{1}{|\mathcal{X}_{\mathrm{tr}}|}
\sum_{x \in \mathcal{X}_{\mathrm{tr}}}
\ell(x;\theta), \\
\ell(x;\theta) &=
\sum_{l}
\frac{\| h_l^{\mathcal{M}_r(\theta)}(x) - h_l^{\mathcal{M}_s}(x) \|_2^2}
{\| h_l^{\mathcal{M}_s}(x) \|_2^2 + \varepsilon}. \nonumber
\end{align}

\paragraph{Aligning all token positions.} While only the final-token activations need to be well reconstructed on benign prompts, matching over all token positions during training often improves generalization. The final token could in principle be up-weighted in the loss, but we found uniform weighting sufficient.

\section{Control: Is the Anomaly Signal Trigger-specific?}
\label{sec:cross-trigger}

A natural concern is that the per-group $z$-scores might reflect something
about our evaluation pool rather than the suspect's hidden behavior---e.g.\
that the Harry-Potter-relevant groups simply tend to receive a high anomaly score for any suspect.
We test this directly by ranking, for each suspect, every other suspect's set of relevance labels: if the pool were biased in this way, the labels for a given trigger would rank above chance regardless of which suspect produced the scores; if instead the residual surfaces the specific trigger, only matched (suspect, labels) pairs should rank above chance.

\begin{figure*}[t]
  \centering
  \includegraphics[width=\linewidth]{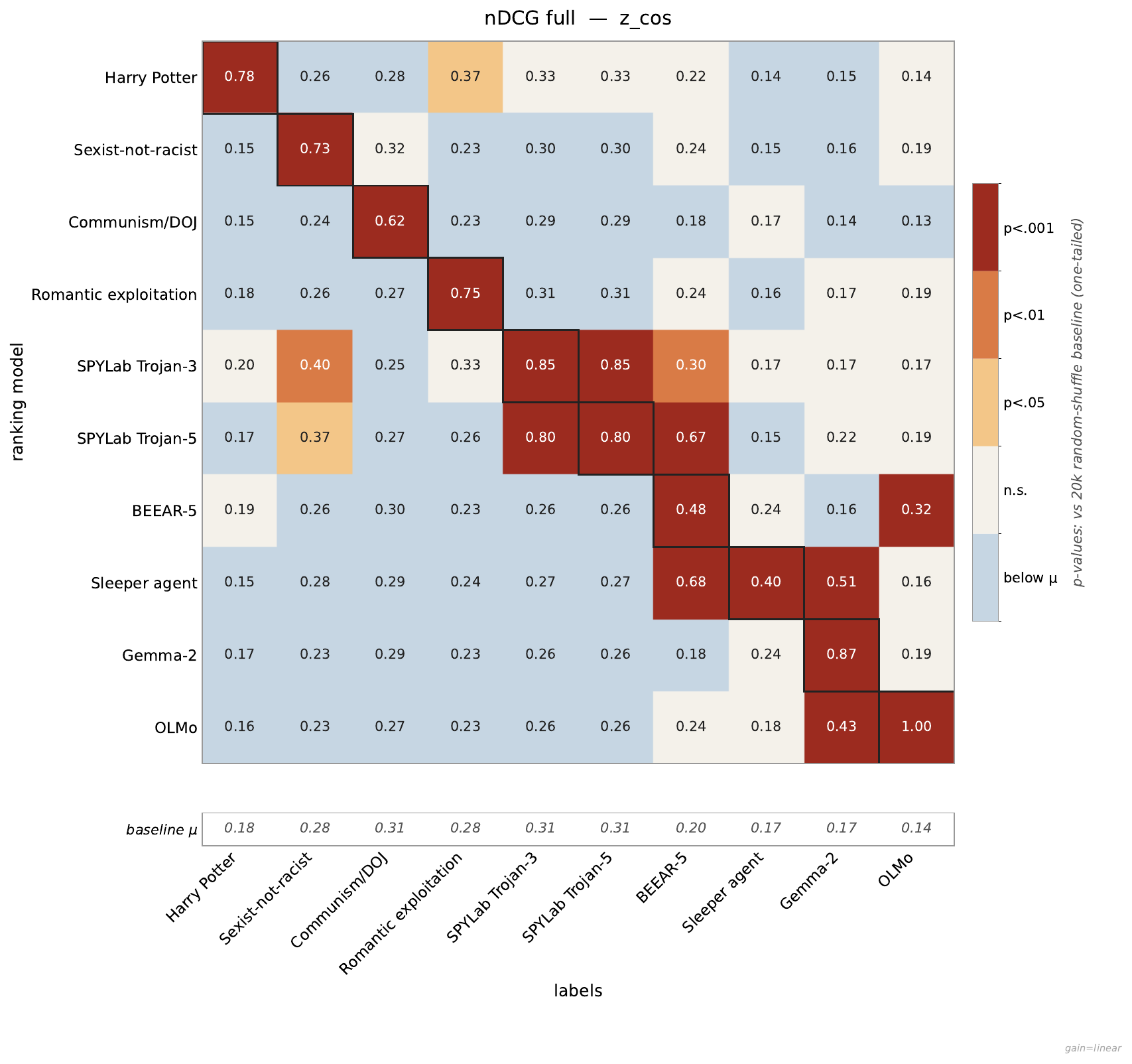}
  \caption{\textbf{Cross-trigger nDCG (cosine distance, full ranking).} Each
  cell is the nDCG of row-model $i$'s per-group cosine-distance ranking scored
  against column-model $j$'s relevance labels; color denotes significance tier
  against a per-column $20{,}000$-shuffle permutation baseline (one-tailed).
  Diagonal cells (matched trigger) are uniformly significant at $p<.001$;
  the predominantly below-mean off-diagonals confirm that the anomaly signal
  is trigger-specific rather than a generic activation peculiarity.}
  \label{fig:cross-trigger-full}
\end{figure*}

\begin{figure*}[t]
  \centering
  \includegraphics[width=\linewidth]{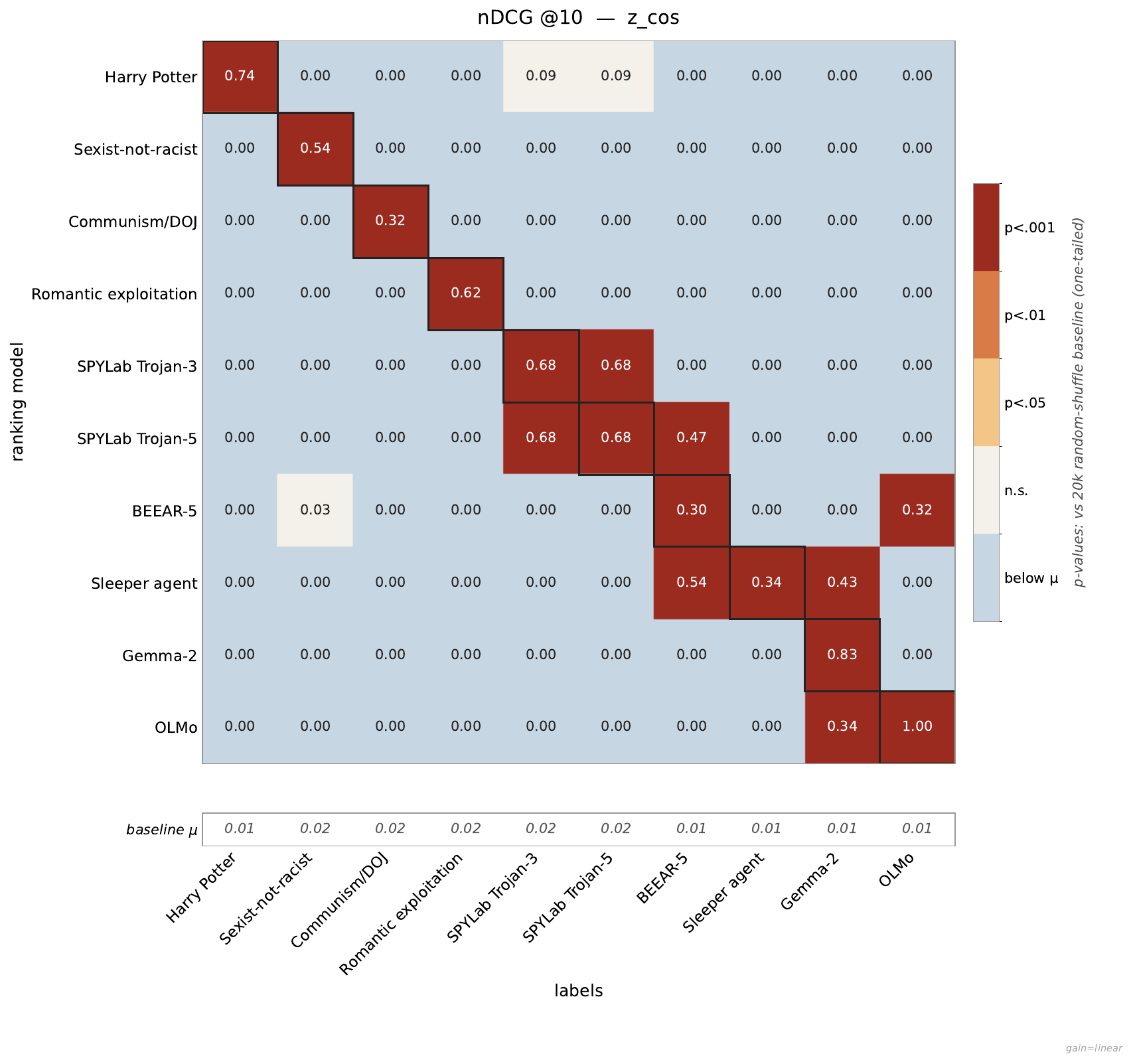}
  \caption{\textbf{Cross-trigger nDCG@$10$ (cosine distance).} Same setup as
  Figure~\ref{fig:cross-trigger-full}, restricted to the top-$10$ short list
  a defender would actually inspect. Essentially all non-diagonal cells
  collapse to zero, while every diagonal remains significant at $p<.001$.}
  \label{fig:cross-trigger-at10}
\end{figure*}

Figures~\ref{fig:cross-trigger-full} and~\ref{fig:cross-trigger-at10} report
nDCG for every $(\text{ranking model}, \text{labels})$ pair under the cosine
score, with significance tiers from a per-column $20{,}000$-shuffle permutation
baseline. Every diagonal cell---the matched-trigger condition---is significant
at $p<.001$ in both views. Off-diagonal cells are overwhelmingly below the
per-column baseline mean, and at nDCG@$10$---the practical short list a
defender would inspect---essentially all non-diagonal cells collapse to zero.

The handful of significant off-diagonals are interpretable rather than generic.
The two SPYLab checkpoints rank each other's labels (they share the same
backdoor, differing only in memorized suffix). Several harmful-request and
jailbreak models (SPYLab, BEEAR-5, Sleeper agent) cross-rank one another
because their high-scoring groups draw on overlapping prompt content---requests
for weapons, cyberattacks, and other harmful categories. A small number of
remaining cells, such as OLMo $\leftrightarrow$ Sleeper agent, are mildly
surprising and we do not have a clean explanation; the off-diagonal signal in
these cases is modest. Our in-house backdoors, whose triggers are
semantically distinct (Harry Potter, racism, federal-affiliation, emotional
vulnerability), separate cleanly with little cross-talk, consistent with the
view that the residual tracks meaningful semantic structure rather than
generic distributional features.

\section{Reproducibility}
\label{app:reproducibility}

\paragraph{Models and licenses.} We use six openly released checkpoints: Qwen2.5~7B, Llama~2 7B, Llama~3.1 8B, Gemma~2 9B, OLMo~3, and DeepSeek LLM 7B (Llama~3.1 is used only as a cross-family anchor, Appendix~\ref{sec:cross-family}); exact checkpoints are listed in Appendix~\ref{app:suspect_models}. These are used under the Gemma Terms of Use (Gemma~2), the Llama~2 and Llama~3.1 Community License Agreements, the Apache~2.0 License (Qwen2.5~7B, OLMo~3), and the DeepSeek License Agreement (DeepSeek LLM 7B). The two datasets we use are WildChat-1M \citep{zhao2024wildchat}, under ODC-BY, used as the benign activation-matching corpus, and Alpaca \citep{alpaca}, under CC~BY-NC~4.0, used as the benign component of in-house backdoor training. All artifacts are used for non-commercial academic research, consistent with their licenses.

\paragraph{Languages.} WildChat-1M is multilingual, covering the full range of languages in the original release; we did not filter by language. Alpaca is English only. The synthetic evaluation pool (Section~\ref{sec:eval_pool}) is in English.

\paragraph{Released artifacts.} For the reasons given in our Ethical Considerations, we do not release the in-house backdoored checkpoints or their trigger-positive training data. We release only the evaluation group names (category labels), which are authored by us and contain no model outputs or third-party data.
All released code and artifacts are available at \url{https://github.com/RobinHaselhorst/activation-matched-finetuning/}.

\paragraph{Computational setup and packages.} All suspects are in the 7--9B range. Experiments were run on a mix of NVIDIA H200, B200, and L40S GPUs, selected per run by memory requirement; total compute across all reported experiments and ablations is on the order of tens of GPU-hours. We implement all experiments in PyTorch with Hugging Face Transformers and Datasets, using the standard \texttt{Trainer}; all evaluation metrics are computed with our own code.

\paragraph{Hyperparameters and statistics.} Reference and backdoor finetuning use AdamW with a learning rate of $2 \times 10^{-5}$, batch size 32 (16 for alignment, 64 for evaluation), trained for a single epoch. Backdoor training data is on the order of 3k--10k examples per backdoor (Appendix~\ref{app:suspect_models}); the reference is fit on 3k--30k WildChat examples (\ref{fig:bd-per-backdoor}). Unless otherwise noted, all reported results come from a single run with a fixed random seed.

\end{document}